\documentclass{article}
\usepackage[utf8]{inputenc}
\usepackage{authblk}

\title{NNN: \underline{N}ext-Generation \underline{N}eural \underline{N}etworks for Marketing Measurement}
\author[]{Thomas Mulc}
\author[]{Mike Anderson}
\author[]{Paul Cubre}
\author[]{Huikun Zhang}
\author[]{Ivy Liu}
\author[]{Saket Kumar}
\affil[]{Google}
\date{June 2025}

\usepackage{booktabs}
\usepackage{graphicx}
\usepackage{amsmath}
\usepackage{amsfonts}
\usepackage{float}
\usepackage{subcaption}
\usepackage{hyperref}
\usepackage[margin=1in]{geometry} % Set all margins to 1 inch

\usepackage{tikz}
\def\checkmark{\tikz\fill[scale=0.4](0,.35) -- (.25,0) -- (1,.7) -- (.25,.15) -- cycle;} 

\usepackage{xcolor}
\usepackage{listings}

\lstdefinestyle{mystyle}{
    backgroundcolor=\color{gray!10},   % Light gray background
    commentstyle=\color{green!60!black},
    keywordstyle=\color{blue},          % Keywords in blue
    numberstyle=\tiny\color{gray},    % Line numbers smaller and gray
    stringstyle=\color{purple},        % Strings in purple
    basicstyle=\ttfamily\scriptsize, % Use typewriter font, footnote size
    breakatwhitespace=false,
    breaklines=true,                 % Enable automatic line breaking
    captionpos=b,                    % Put captions below the code
    keepspaces=true,
    numbers=none,                    % Show line numbers on the left
    numbersep=5pt,                   % Space between numbers and code
    showspaces=false,                % Don't show visible spaces
    showstringspaces=false,          % Don't show visible spaces in strings
    showtabs=false,                  % Don't show visible tabs
    tabsize=2,                       % Set tab width to 2 spaces
    frame=single,                    % Add a frame around code blocks
    rulecolor=\color{black!30},      % Color of the frame
    title=\lstname                   % Show filename above if provided
}
\begin{document}
\maketitle

\section{Abstract}

We present NNN, an experimental Transformer-based neural network approach to marketing measurement. Unlike Marketing Mix Models (MMMs) which rely on scalar inputs and parametric decay functions, NNN uses rich embeddings to capture both quantitative and qualitative aspects of marketing and organic channels (e.g., search queries, ad creatives). This, combined with its attention mechanism, potentially enables NNN to model complex interactions, capture long-term effects, and improve sales attribution accuracy. We show that L1 regularization permits the use of such expressive models in typical data-constrained settings. Evaluating NNN on simulated and real-world data demonstrates its efficacy, particularly through considerable improvement in predictive power. In addition to marketing measurement, the NNN framework can provide valuable, complementary insights through model probing, such as evaluating keyword or creative effectiveness.

\section{Introduction}
In today's data-driven marketing landscape, accurately attributing sales to specific campaigns remains a paramount challenge. Marketers grapple with quantifying the return on investment for their marketing spend, seeking reliable estimates of the sales uplift generated by each channel. While randomized controlled experiments offer causal rigor, they are often impractical or prohibitively expensive to implement across a full media mix and typically cannot predict effectiveness at untested spend levels.

Marketing Mix Models (MMMs), with roots tracing back to the 1950s \cite{gujar2024evolution, borden1964concept}, have seen a resurgence in recent years \cite{chan2017challenges, jin2017bayesian, lightweight_mmmgithub, robyn}, offering a means to estimate the effectiveness of diverse media channels using aggregated, observational data. This renewed interest is partly fueled by the increasing difficulty of long-term tracking for attribution and user-level experiments \cite{ravid2025marketing}, especially given the deprecation of third-party cookies and stricter privacy regulations like GDPR, CCPA \cite{california_consumer_privacy_act} and Apple's Intelligent Tracking Prevention (ITP) \cite{apple_safari_itp}. MMMs circumvent these tracking needs by operating on aggregated data (e.g., weekly impressions and sales per region).

However, current MMM methods face well-documented limitations \cite{chan2017challenges,runge2024packaging}.  First, MMM datasets are often limited in  samples relative to the number of factors influencing sales, which can lead to statistically overdetermined models \cite{chan2017challenges} which has historically limited the feasibility of employing more complex, data-hungry techniques like deep neural networks. Second, the common assumption\footnote{There are a few exceptions such as \cite{uber2021}.} of time-invariant response curves makes it challenging to account for variations in marketing effectiveness due to changes in ad creative quality, targeting strategies, or external factors like seasonality\footnote{Such as changes in effectiveness during the holiday shopping season.}. Finally, capturing long-term brand effects is notoriously difficult; parametric adstock functions typically used to model carryover effects often lose identifiability from the baseline beyond a few months.

\begin{figure}[H]
    \centering
    \includegraphics[scale=.30]{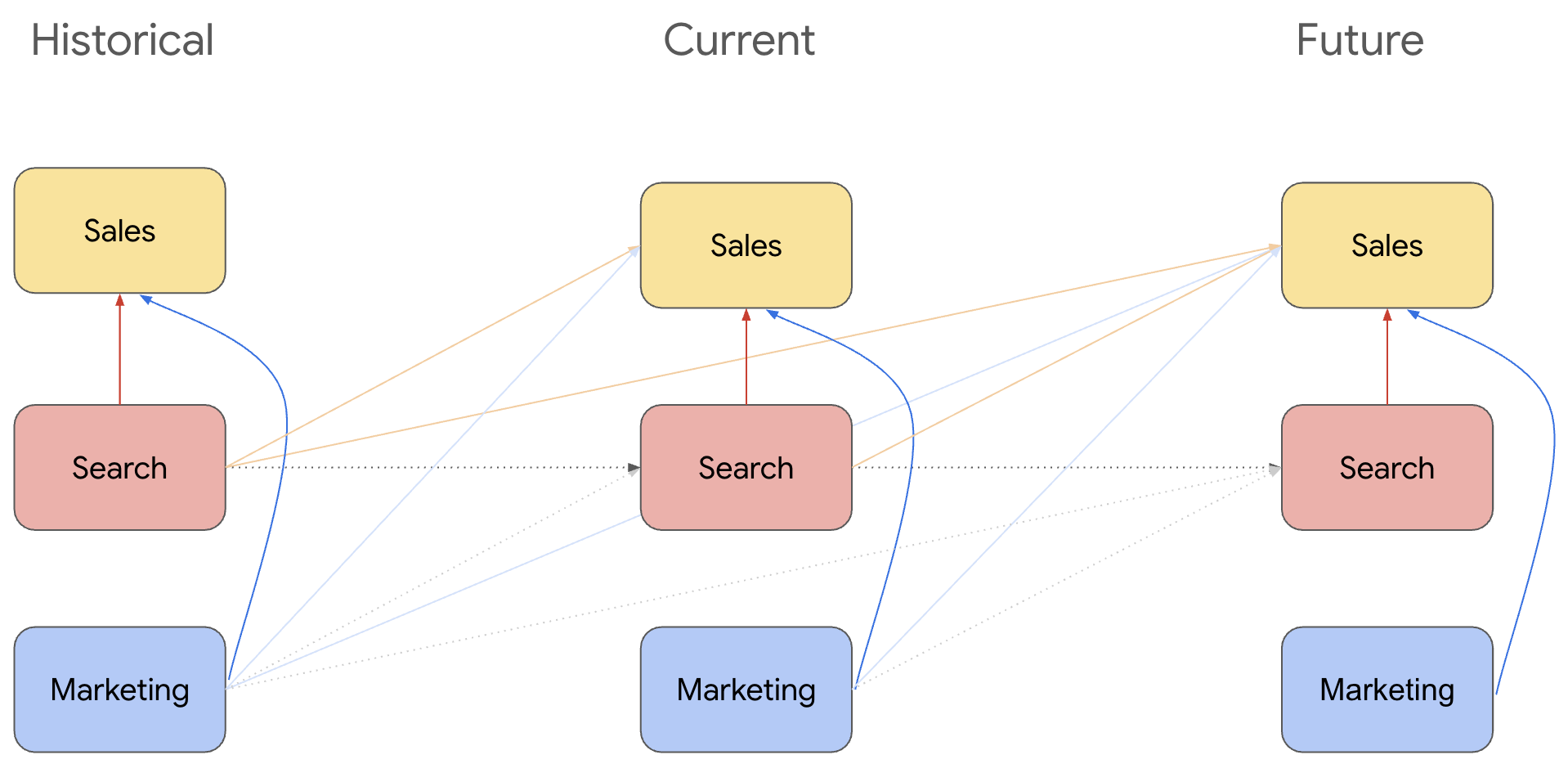}
    \caption{High-level causal structure in our models. Red arrows represent the links from organic Search to sales, modeled in \cite{mulc2024compressing}. Yellow arrows represent new links from previous search to current sales, which are implemented through our Transformer. Blue arrows represent the new causal links between marketing and sales. Dashed arrows represent the links the search predictive component of the model. Note that this component is autoregressive, whereas the sales is not. More complicated causal structures are possible with our framework as well, but mostly outside the scope of this work.
}
    \label{fig:causal}
\end{figure}

To address these challenges, this paper introduces NNN (Next-Generation Neural Networks for Marketing Measurement), a novel paradigm leveraging recent advancements in deep learning and representation learning. This is an experimental approach which uses a different foundational modeling method compared to current MMM methodology. Some key differences include:

\begin{itemize}
    \item \textbf{Rich Data Representation.} Instead of relying on scalar inputs (e.g., spend), NNN utilizes high-dimensional embeddings to represent both marketing activities (e.g., Search Ads, YouTube Ads) and crucial organic signals (e.g., Google Search queries). These embeddings capture not only quantitative volume but also qualitative aspects like ad creative content or search query semantics, allowing the model to directly address the challenge of time-varying creative effectiveness.
    \item \textbf{Modeling Long-Term Effects via Intermediary Signals.} NNN explicitly models the influence of marketing activities on intermediary signals highly predictive of consumer behavior, specifically Google Search query patterns (as depicted in Figure \ref{fig:causal}). This, combined with a Transformer architecture capable of attending to long historical sequences, provides a mechanism for capturing marketing effects that unfold over extended time horizons, potentially overcoming the limitations of standard adstock models.
    \item \textbf{Neural Network Power with Regularization.} The framework employs neural networks and a custom Transformer to learn complex non-linear interactions between channels and over time. We demonstrate that L1 regularization makes it feasible to train these highly parameterized models effectively, even within the often data-constrained settings typical of MMM, mitigating concerns about overfitting.
    \item \textbf{Objective Model Selection.} NNN facilitates model selection based on rigorous evaluation of predictive performance (e.g., MAPE, R$^2$) across multiple held-out datasets, particularly temporally distinct test sets, providing a robust alternative or complement to traditional criteria based on parameter plausibility.
\end{itemize}

Our paper is structured as follows. We give an overview of related working influencing our approach (Related Work). Next, we discuss the data used (Datasets), and how they are formatted in our method (Data Structure). We give an overview of the computation of the model (Model Overview), then go into further details regarding the how traditional data is augmented and embedded (Input Data Representations). We discuss the intricacies of our Transformer (Transformer) and details about how the model makes the final predictions (Prediction Heads). We discuss training details (Model Training) and attribution methods (Sales Attribution). We give an overview of our experiments and then discuss the results and analysis (Experiments). Finally, we give a discussion about how our method compares to existing measurement approaches (Discussion), and give a concluding overview (Conclusion).

% (Keep the figure environment)

\section{Related Work}
Our proposed NNN framework builds upon and diverges from several strands of research in marketing measurement, primarily Media Mix Modeling, the application of neural networks in this domain, and the use of embeddings to represent media signals. This section situates NNN within this existing literature.

\subsection{Media Mix Modeling Evolution}
The field of Media Mix Modeling (MMM) provides the foundational context for NNN, offering established methods for assessing advertising ROI using historical, aggregate data \cite{gujar2024evolution, borden1964concept}. Recognizing limitations in early linear models, Bayesian Media Mix Modeling (BMMM) emerged, introducing techniques to model ad carryover and saturation (shape effects) more flexibly, often using Hill and adstock functions \cite{ jin2017bayesian}. BMMM leverages Bayesian inference (e.g., MCMC methods) to incorporate prior knowledge into model estimation, a crucial aspect given the often data-constrained nature of MMM problems. This Bayesian framework was further extended to handle geographical variations through hierarchical modeling \cite{sun2017geo}. Meridian, a Google Bayesian MMM framework integrated advancements of MMM research over the past decade. It incorporates features like ROI priors and calibration \cite{zhang2024bayesian}, reach and frequency data integration \cite{zhang2023reach}, and paid search modeling. While these approaches are powerful and innovative, here we are exploring an entirely new research direction that does not rely on scalar inputs or pre-defined functional forms for temporal effects.

\subsection{Neural Networks in Media Mix Modeling}
The potential for neural networks (NNs) in MMM has been recognized for some time \cite{tedesco1998neural}, owing to their ability to capture complex non-linearities. However, concerns over interpretability and the data requirements for training large NNs have historically limited their adoption in practice. Recent efforts aim to bridge this gap. For instance, CausalMMM \cite{gong2024causalmmm} attempts to learn causal structures using Granger causality \cite{granger1969investigating} and employs a variational autoencoder (VAE) \cite{doersch2016tutorial} with a dedicated causal relational encoder and a marketing response decoder where the temporal response is learned alongside the saturation response. While innovative in its use of NNs and causality, this approach still operated on scalar representations of media activity. NNN builds on the potential of NNs but focuses on leveraging richer input representations via embeddings.

\subsection{Embedding-Based Media Representations}
A core innovation of NNN is its use of high-dimensional embeddings, drawing heavily on the work presented in \cite{mulc2024compressing}. That work demonstrated that organic signals, specifically aggregated Google Search query embeddings, could be highly predictive of real-world outcomes like flu rates and auto sales. This contrasted with earlier studies that used Google Search data but often relied on counts of specific keywords, necessitating significant feature engineering \cite{cook2011assessing, lazer2014parable, lampos2015advances, Varian2009, Woloszko2020}. The key insight from \cite{mulc2024compressing} was using pre-trained language models to embed search terms and then aggregating these embeddings (via summation in SLaM) to create a dense ``search embedding'' for a given time and geo, capturing both the \textit{volume} and \textit{semantic content} of search activity as a proxy for organic consumer demand.  In our work, we show that the search embedding is a highly effective baseline for regression-based measurement models, and we introduce additional embeddings for other paid marketing channels such as Google Search ads and YouTube ads.

\subsection{Terminology and Conventions}

Throughout this paper, we adopt specific terminology for clarity. We use ``MMM'' to refer generally to current Marketing Mix Modeling methods, which could be Bayesian or Frequentist, and generally use impressions or other scalar quantities to represent media. In contrast, ``NNN'' denotes our proposed neural network-based methodology which uses embeddings and scalars to represent media.

While our framework is adaptable, this work primarily focuses on predicting business ``sales'' as the main response variable or Key Performance Indicator (KPI), similar to how U.S. auto registrations or influenza rates were used as targets in related work \cite{mulc2024compressing}. Likewise, although other organic channels could potentially be incorporated or substituted, we utilize Google Search (``Search'') as the principal organic channel analyzed and predicted within NNN. Consequently, we may use the terms ``Search'' and ``organic channel'' interchangeably when referring to this specific input or prediction target.

A glossary defining the mathematical notation used (e.g., for input tensors, dimensions, parameters) and commonly used terms can be found in Appendix \ref{sec:glossary}.

\section{Datasets}
We experiment on three different datasets that span both simulated data and real-world data that includes marketing spend and a revenue / sales figure.

\begin{figure}
    \centering
    \includegraphics[scale=.30]{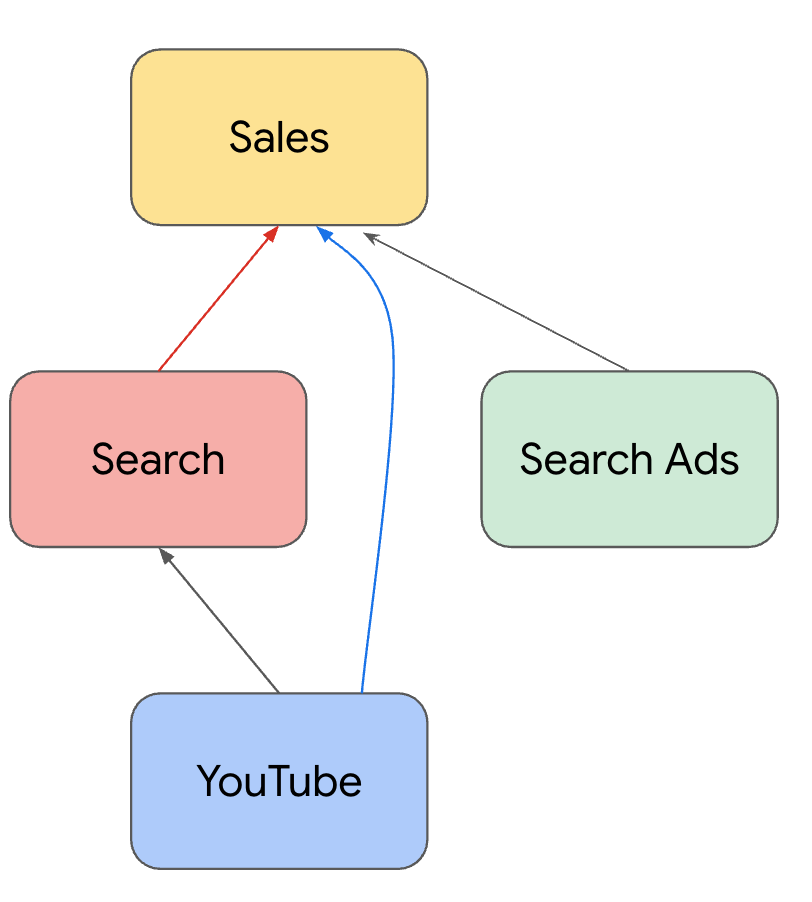}
    \caption{Simulated Data DAG.}
    \label{fig:sim_dag}
\end{figure}

\begin{itemize}
    \item Simulated data.  We simulate data according to the Directed Acyclic Graph (DAG) in Figure \ref{fig:sim_dag}.  This is a highly oversimplified DAG (it lacks an arrow from search to search ads, for example, which is known to be an important channel in correcting for bias in paid search advertising; see \cite{chen2018bias}). For our initial explorations in this paper, however, we wanted instead to focus on the route from YouTube to Search to Sales. This pathway cannot usually be assessed with single-model MMMs since Search counts as a mediator in this framework, but our NNN framework with KPIs for both Search and Sales allows us to infer both direct (YouTube to Sales) and indirect (YouTube to Search to Sales) components of the effect of YouTube. We use this simplified DAG to illustrate a potential application of our model framework, but do not intend it to be a comprehensive DAG reflective of a full marketing mix.
    
    Data was randomly generated using the \texttt{test\_utils} module of the Meridian package to simulate 130 weeks of data over 100 random geos. We used this module for simulated data generation given its ease of use, but no properties of the simulated data are specific to the package. We created two channels of random media data (YouTube and Search Ads) and one channel of random controls data (Search), applying adstock and Hill functions to the media channels. In order to model sales driven directly by YouTube, we assumed an adstock retention rate $\alpha_m = 0.75$ and a Hill function effective concentration $e_c = 1.0$ and applied these functions to the random YouTube media data with arbitrary normalization. For the indirect sales driven by YouTube, we applied a different set of adstock and Hill functions to the YouTube media data ($\alpha_m = 0.5$ and $e_c = 3$) and added those normalized impressions to the random search data so that they contributed an additional 20\% or so on top of the random search data. We then generated sales from the sum of the random search impressions and the YouTube-generated search impressions, assuming a flat conversion rate of 0.1\%. Finally, search ads (which in this simplified model are not causally connected to search) also generated sales using $\alpha_m = 0.3$ and $e_c = 1$. 
    
    We introduced a new complication into the simulated dataset by adding a multiplicative term to all four pathways (YouTube to Search, YouTube to Sales, Search Ads to Sales, and Search to Sales) with uniformly distributed random values (varying over time but not geo) centered around 1. This is intended to mimic the ``intent'' described in \cite{mulc2024compressing}, alongside the ``volume'' described in that paper which is represented by the media impressions data. It may be easiest to think of this as a measure of creative effectiveness for the marketing channels in the model, in which case we are explicitly allowing creative effectiveness to vary at a weekly level in the model. In order to give the NNN a chance to measure this effect, we also generate a random 256-dimensional embedding representation of the creative effectiveness for every week by interpolating linearly between two random embeddings corresponding to the highest possible creative effectiveness and the lowest possible creative effectiveness in our model.

    In the end we have two versions of this dataset: High variance and low variance, differing only in the magnitude of the random noise corresponding to ``intent'' or creative effectiveness. For the high-variance dataset, this varies uniformly between 0.5 and 1.5, while in the low-variance dataset it varies uniformly between 0.8 and 1.2.  We compare our method against a baseline built using current MMM methods on these datasets on both sales-prediction error-rate evaluation and on sales channel attribution estimates, because the ground-truth attributions are known when creating the data.
    \item Real-world, Regional-level.  (4 regions).  This data is real-world marketing and sales data, aggregated at a regional- and weekly-level.  We use it for an apples-to-apples comparison of NNN against a baseline MMM in a real-world setting where priors have already been established.  We use only the marketing volume as model inputs (quantitative information) in both NNN and the baseline MMM.  The Search embeddings are generated using approximately 8M queries that are categorically related to the business's sales.
    \item Real-world, DMA-level (210 DMAs\footnote{Designated Marketing Areas}).  This is an additional real-world dataset that uses more granular geo-regions from a different time period, but it is also aggregated at the weekly level.  We use our embedded version of marketing channels (Search Ads and YouTube) in these models (i.e., we use both quantitative and qualitative information).
\end{itemize}

\section{Data Structure}
In NNN, we utilize a rank-4 tensor, $X$, with shape $(G, T, C, D)$ to represent all target variables, organic channels, and media channels for all geos across all time-steps.  $G$ is the number of geography (geo) regions; $T$ is the total number of time-steps; for example, if we have 52 weeks of weekly data, then $T=52$;  $C$ is the total number of channels, including all media channels, organic channels (e.g., Search), and the target channel (e.g. sales);  $D$ is the maximum size of the embedding dimension across all channels. Although sales--and historically most media channels--are represented as scalar quantities, we utilize embeddings from foundational models such as language models, and therefore, $D$ tends to be a few hundred of dimensions, at least.

We use NumPy notation to index our tensor.  For example, $X_{:,t,c,:} \in \mathbb{R}^{G \times D}$ is a tensor that represents all geos and embedding dimensions at time $t$ for channel $c$.The time dimension assumes uniform temporal spacing, with the index increasing monotonically (e.g., for weekly data, $t=0$ is week 1, $t=1$ = week 2, ...).

For scalar quantities, such as sales, we pad the scalar to $D$-dimensional vector.  For example, $X_{g,t,\text{sales},0}$ contains the scalar of sales for the time $t$ and geo $g$, but $X_{g,t,\text{sales},1:}$ is a vector of zeros.  Similarly, any channel that is not a scalar but is instead a vector whose dimensionality is less than $D$ is also padded with zeros to be of length $D$.  This allows us to handle embeddings from multiple models that may not use the same embedding size.

\section{Model Overview}
We will model the input data $X$ using a neural network, $F$ with parameters $\Theta$.  We use a neural network because it gives us the flexibility \cite{LeCun2018DifferentiableProgramming, blondel2024elements, goodfellow} to model the causal dependencies in Figure \ref{fig:causal} via functional composition that is easily differentiable with modern software packages like JAX \cite{jax2018github}.

Our baseline NNN model is based on the Search-to-Sales dependency in Figure \ref{fig:causal}.  Previous work has shown that Google Search is a good proxy for the consumer demand, and even captures seasonal effects; we use an embedded version of Search where user queries are aggregated according to their semantic representations from a language model \cite{mulc2024compressing}.  We pass this embedded representation of search to a function that is similar to the hill functions found in many modern MMMs to get our estimate of Sales.  We enrich our Sales estimate by including marketing-to-search effects (the blue lines from Marketing to Sales in \ref{fig:causal}) in a similar fashion, utilizing embedded versions of the paid media.  To include the temporal effects in Figure \ref{fig:causal} (the yellow lines from Search to Sales and the blue lines from Marketing to Sales), we utilize a Transformer architecture that is able to attend to channels' histories.  This allows the feature representation used to predict sales to depend not just on the current time-period's Search, but also it's history.  Similarly, this is done for all other channels. Finally, our model is allowed to be trained not just to predict sales, but to predict organic channels like Search (the grey lines in Figure \ref{fig:causal}).  This allows our model to capture marketing effects that happen through an intermediary such as Google Search (E.g., Imagine a channel that affects Search, and which then drives sales).

The following sections go into details about each of the components.  For clarity, we provide pseudo code for much of the model computation in the Appendix.

\section{Input Data Representations}

\subsection{Embeddings as Media Representations}
One of the strengths of the MMM framework is that it does not require user-level data in the model, making data collection a tractable challenge for most advertisers even over time baselines of several years. Traditionally marketing models represent media on a given day / week / month as a scalar quantity, as either the volume or spend on a given channel.  This is partly motivated by the levers markets have at their disposal: they can typically increase or decrease the marketing volume. 

Another important aspect of media is not just its quantity, but its content (i.e., quality). This is typically not addressed in MMMs, but we argue that this can also be measured in an aggregate form and incorporated into models of media effectiveness. This type of data is also a potential lever for marketers (e.g., marketers can build new ads with different creatives) and contains rich information about a consumer's willingness to purchase.

We represent media as a $D$-dimensional vector embedding whose magnitude represents the volume and whose direction represents the qualitative components of the media.  For each type of media, a method for generating the direction of it's embedding must be determined, usually by some pretrained foundational model.  Additionally, the metric to use for the magnitude is also channel / embedding specific, although it is usually the advertising impressions.

\subsection{Search Embeddings}
For each user search, we embed the search term, $s$, according to the pretrained Multi Lingual Sentence Encoder (MLSE) \cite{yang2019multilingual} to get a 512 dimensional embedding.  Then following the SLaM procedure \cite{mulc2024compressing}, for each geo and time-period, we generate a search embedding as
$$X_{g,t,\text{Search}} = \sum_{s \in S_{g,t}} e(s)$$
where $t$ represents the time-period, $g$ is the geo, $e$ is MLSE, and $S_{g,t}$ is the set of all searches that happen during time $t$ in geo $g$.  Note that we found $||X_{t,g, \text{Search}}||$ to be highly correlated with $V_{g,t}$, the number of user queries during week $t$ in geo $g$, and thus we use the norm as a convenient substitute for the actual volume during modeling (See \ref{fig:norm_volume} in Appendix for more details).  In our experiments, we include about 8M unique search terms that are categorically related to the sales metric.

\begin{figure}
    \centering
    \includegraphics[scale=.35]{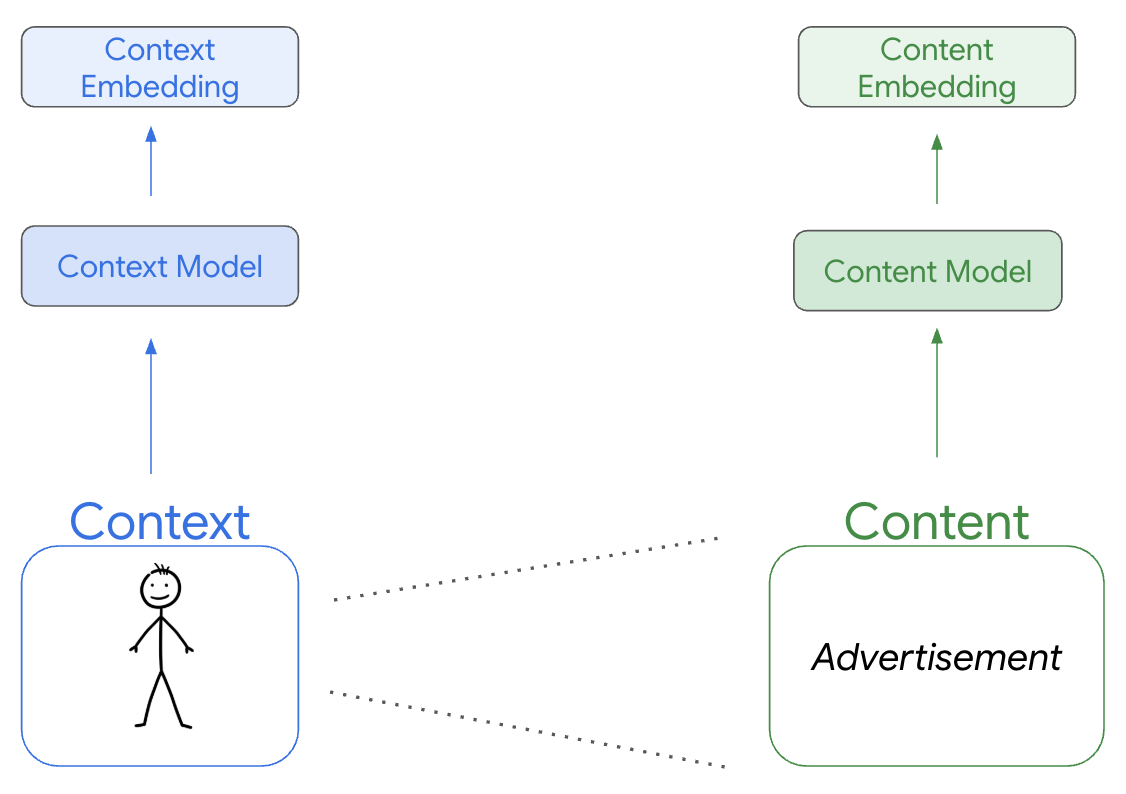}
    \caption{Pictorial representation of an advertisement event.  A member of some marketing audience exists within a certain context (e.g., “A user entered a query ‘X’ into a search engine), represented by the context embedding, is exposed to some advertisement content represented by content embedding.  In our work, we use a language model to represent search queries, which give use context embeddings for Search and Search Ads channels, while we use a video model to represents the advertisement content in the YouTube channel.  In our work, user-level data is never used, however the user was shown here for clarity on where marketing events typically occur.}
    \label{fig:ad_event}
\end{figure}

\subsection{Search Advertisement Embeddings}
For Search Ads, we had the choice to measure either the context (i.e., what terms the users were searching for when they were shown an ad), or the content (i.e., the ad copy material they were shown).  We chose to measure the \textbf{context}.  For each advertising impression, we embed the user's search term (the query) according to the MLSE to get an embedding table.  Then using SLaM and the advertising impressions as the volume term in the aggregation, we get our Search ads embedding, $X_{t,g,\text{SearchAds},:}$, for each time-period and geo.

\subsection{YouTube Advertisement Embeddings}
Similarly, for YouTube advertisements, we could either embed the video the user was watching when they were shown an ad (i.e., context), or we could embed the creative of the ad they were shown, in this case a video advertisement (i.e., content).  We chose to measure the \textbf{content} of the ad.  For each video ad, we passed it through a pretrained video model to get an embedding table. Then using SLaM and the video views as the volume term in the aggregation, we get our YouTube ad embeddings, $X_{t,g,\text{YouTube},:}$ for each geo and time period.

\subsection{Scalars}
In the case where there is no way to use a model to embed data (e.g., in the scenario where only impressions of a channel are stored, and no other contextual information is stored along side it, perhaps for privacy reasons or legacy logging methods), similar to MMM we use only the scalar quantity in our models.  Our architecture requires all channels to be represented as a $D$-dimensional value, so we pad the scalar with $D - 1$ zero values for the other dimensions.  Note that the magnitude of this vector is still the channel volume, which is consistent with all other channels. In this paper we demonstrate one example of independent media variables modeled this way (the real-world dataset), and of course sales (the dependent variable) is also a scalar in our models. 

\section{NNN Transformer}
To incorporate the effect of channel interactions and long-term advertising effects, we introduce a Transformer layer prior to the prediction heads. Without this component, the our model is simply a multi-channel variant of the CoSMo architecture in \cite{mulc2024compressing}.  Recent work has demonstrated that normalizations are not necessarily critical in the function of Transformers \cite{zhu2025transformers}, and we also find that to be true in our work. 

Formally, we define the data / embeddings after the $n$th Transformer layers as

$$X^{(n)} := \text{Transformer}(X^{(n-1)}, \Theta)$$

where $X^{(0)} := X$ is the input data, which is also treated as the Transformer output in the case where zero Transformer layers are used.

\begin{figure}
    \centering
    \includegraphics[scale=.30]{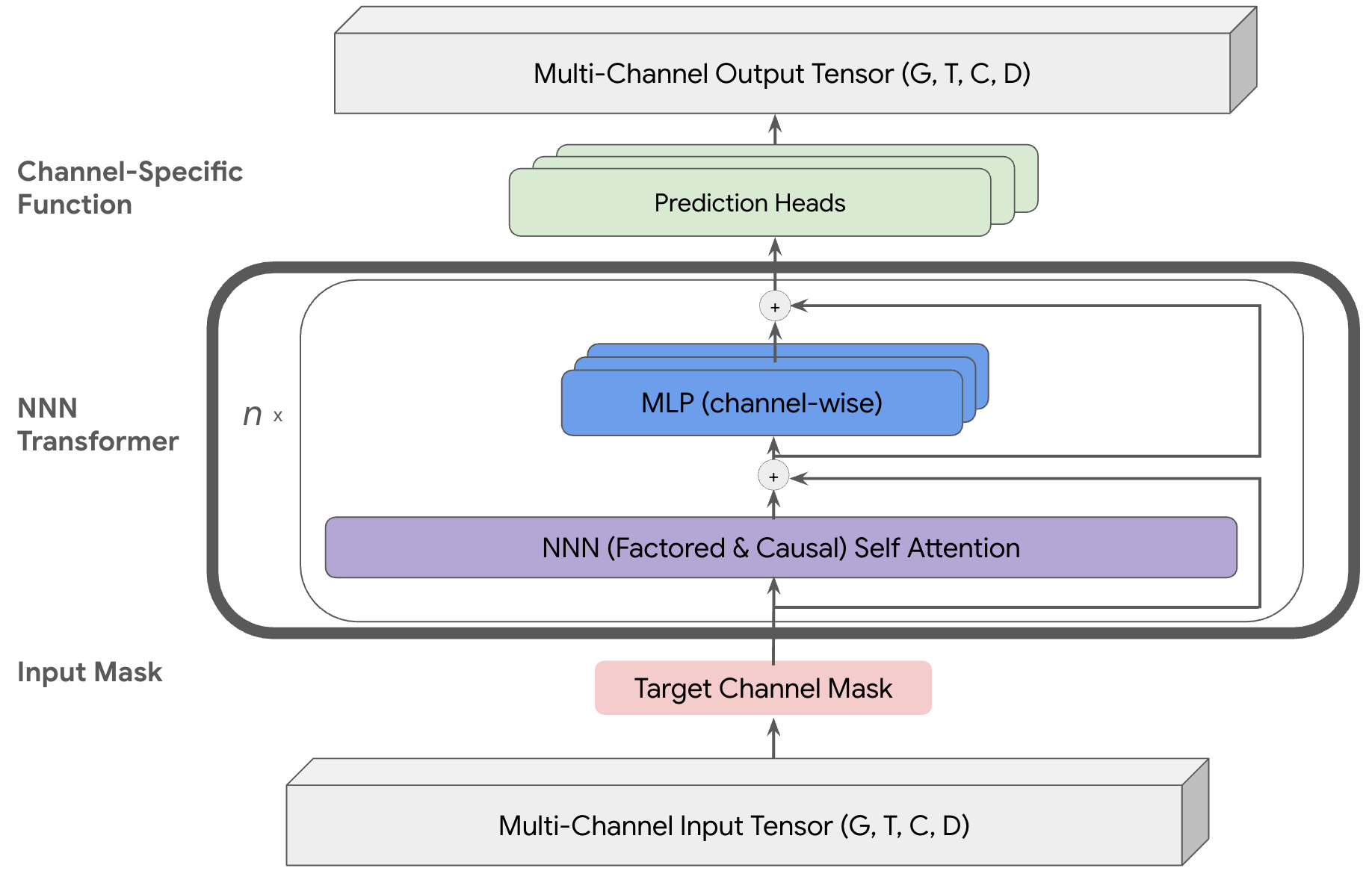}
    \caption{NNN Model Architecture.  The input tensor has shape $(G, T, C, D)$ and the sales targets are masked prior to any other computation.  The masked tensor is then fed into our NNN Transformer $N$ times; if no Transformer layers are used, this component is skipped.  Finally, the intermediate representations of the channels are fed into the prediction heads, where each channel can utilize any subset of the input channels for the final predictions.}
    \label{fig:model}
\end{figure}

Our Transformer is similar to the standard Transformer \cite{vaswani2017attention}, except that we remove the norm components, because the vector magnitudes store valuable information about the quantity of demand / marketing.  Additionally, we do not use additive positional encoding, and instead rely on a custom attention mechanism.  The architecture is shown in Figure \ref{fig:model}.  In each Transformer layer, the data undergoes self-attention and a channel-wise MLP transformation; residual connections are used after both operations.  The Transformer has $n$ layers, after which the data is passed to the prediction heads.

\begin{figure}
    \centering
    \includegraphics[scale=.40]{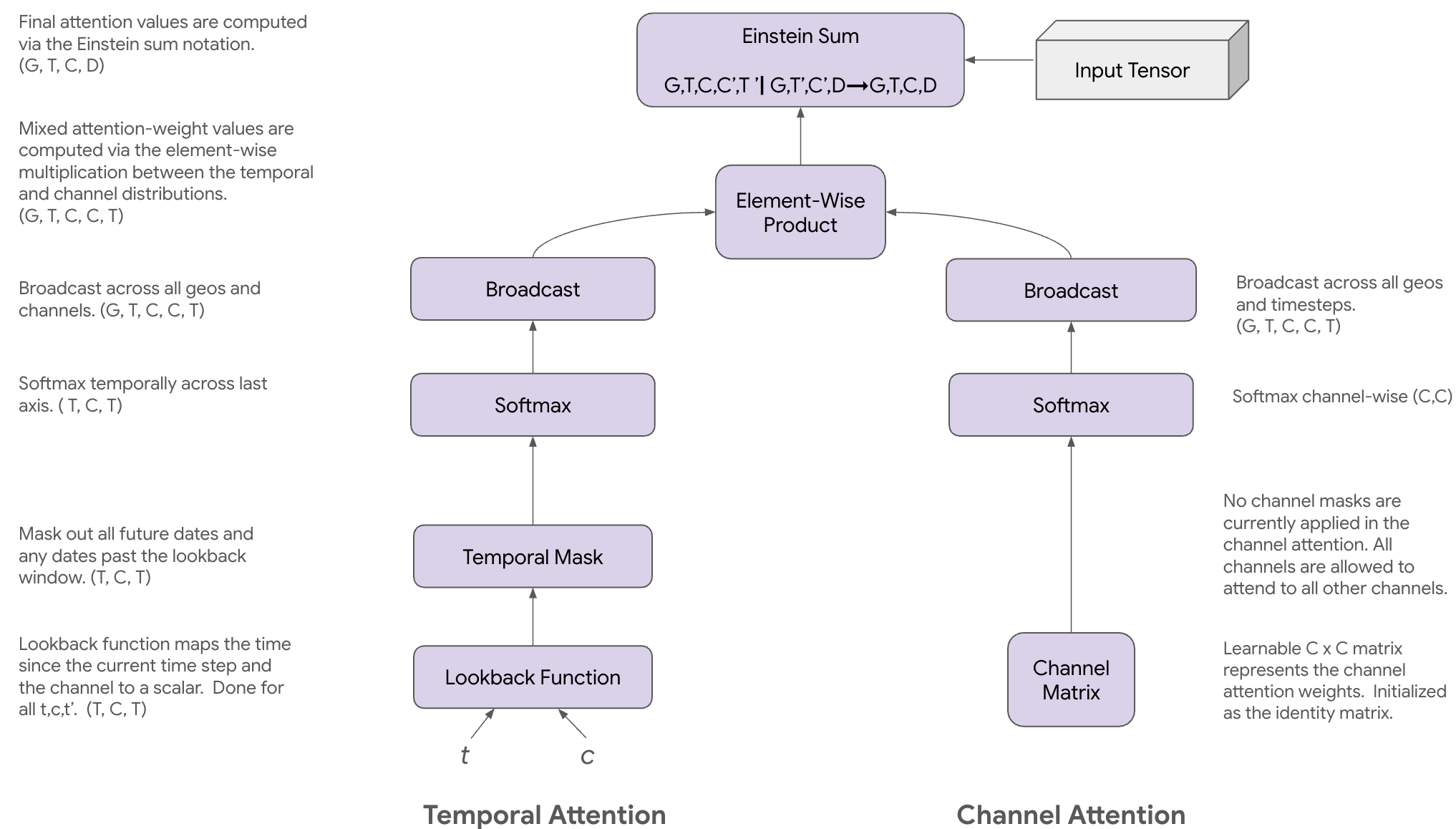}
    \caption{Factored Self-Attention.}
    \label{fig:factored_attn}
\end{figure}

\subsection{Attention}
In our model, we use historical information across marketing channels to predict sales or other business KPIs on a given day $t$. A key innovation in our NNN framework is that we do not have to specify a functional form for the time lag between media exposure and sales. Instead, we employ attention mechanisms \cite{bahdanau2014neural, vaswani2017attention} that learn which channels and prior time steps are most relevant for the current prediction context (channel $c$, time $t$).

We initially experimented with standard Key, Query, Value attention using positional embeddings but encountered several challenges:
\begin{itemize}
    \item Standard absolute positional encodings depend on an arbitrary data start date.
    \item Learning the $T^2$ pairwise interactions for temporal attention can be data-intensive, especially given a typical dataset size of $T \times G$ points.
    \item Extrapolating positional encodings beyond the training time range often leads to poor out-of-distribution performance.
\end{itemize}

To address these issues, we developed a custom \textbf{Factored Self-Attention} mechanism. This approach separates the attention calculation into two distinct components: Temporal Attention and Channel Attention, as illustrated in Figure \ref{fig:factored_attn}. This factorization allows the model to learn temporal dependencies and cross-channel interactions more efficiently.

\subsection{Temporal Attention}
Instead of relying on content-based dot-product attention between keys and queries for temporal relationships, we learn an attention score function $a_{\text{time}}$ that primarily depends on the relative time difference between the query time $t$ and key time $t'$. Crucially, this function can also incorporate the identity of the target channel $c$, allowing different channels to exhibit distinct temporal attention patterns.

\begin{figure}
    \centering
    \includegraphics[scale=.42]{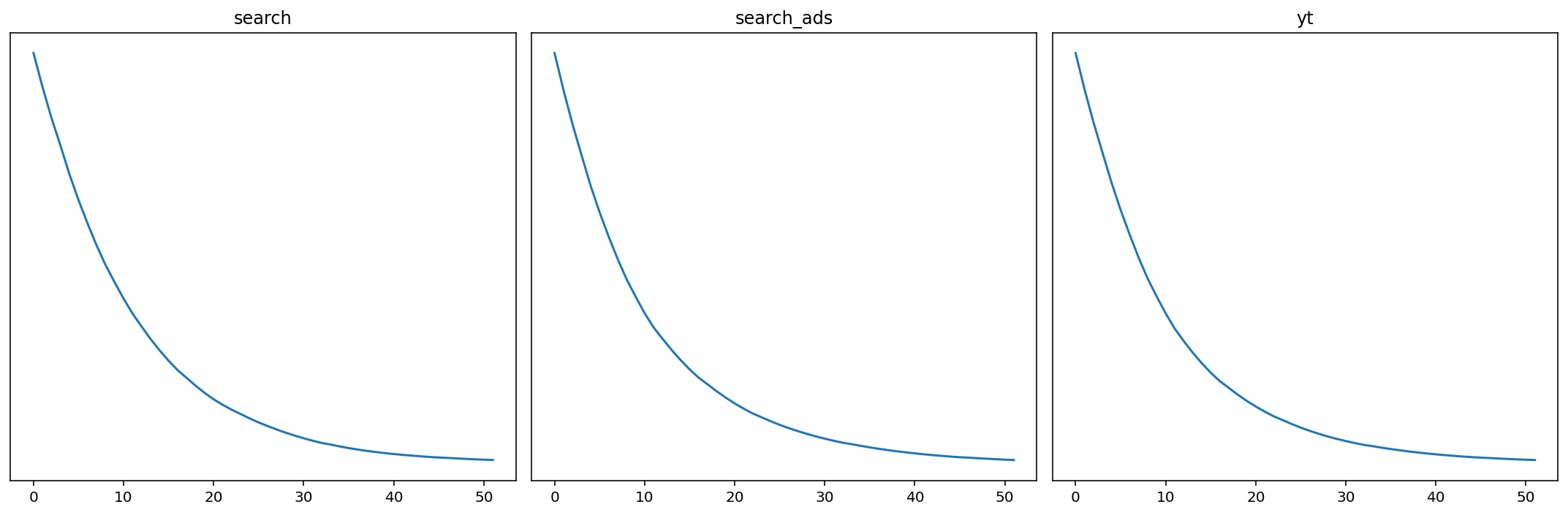}
    \caption{Learned attention for all channels of the first layer of a model trained on Low Variance Simulation; the model learns to focus on nearby search data rather than data farther away.  We mask out all values past 52 weeks.  The function appears to approximate an adstock function, although this structure was never explicitly given in the MLP.}
    \label{fig:learned-attention}
\end{figure}

Let $X \in \mathbb{R}^{G \times T \times C \times D}$ be the input tensor (geo, time, channel, features). Let $\omega$ be the lookback window size.
The input to the attention score function includes:
\begin{enumerate}
    \item The normalized time difference $\Delta_{t, t'} = (t - t') / \omega \in \mathbb{R}$.
    \item A one-hot encoding $E_c \in \mathbb{R}^C$ representing the target channel $c$.
\end{enumerate}
The attention score function $a_{\text{time}}: \mathbb{R}^{1+C} \rightarrow \mathbb{R}$ is implemented as a single-hidden-layer MLP with ReLU activation and 128 hidden units
$$ s_{t, t', c} = a_{\text{time}}(\text{concat}(\Delta_{t, t'}, E_c)).$$
This raw score $s_{t, t', c}$ represents the unnormalized attention from target $(t, c)$ to source time $t'$. We arrange these scores into a tensor $S_{\text{raw}} \in \mathbb{R}^{T \times T \times C}$. (Broadcasting over the geo dimension $G$ is applied in practice).
As shown in Figure \ref{fig:learned-attention}, this formulation allows the model to learn smooth temporal decay functions, emphasizing recent data similarly to an adstock function, without relying on the input sequence values $X$, or a modeler to hand-specify the decay.

To ensure causality and limit the attention span, we apply an additive mask $M \in \{0, -\infty\}^{T \times T}$
$$ M_{t, t'} = \begin{cases} 0 & \text{if } t - \omega < t' \le t \\ -\infty & \text{otherwise} \end{cases} .$$
The masked scores $S' \in \mathbb{R}^{G \times T \times C \times T}$ (after broadcasting $S_{\text{raw}}$ and $M$, and transposing) are computed as
$$ (S')_{g, t, c, t'} = (\text{transpose}(S_{\text{raw}}))_{g, t, c, t'} + M_{t, t'} .$$
Finally, the temporal attention weights $W_{\text{time}} \in \mathbb{R}^{G \times T \times C \times T}$ are obtained via softmax over the source time dimension $t'$, scaled by a temperature $\tau$
$$ (W_{\text{time}})_{g, t, c, t'} = \text{softmax}_{t'}\left( \frac{(S')_{g, t, c, :}}{\tau} \right)_{t'} .$$

\subsection{Channel Attention}
To allow information mixing across channels, we introduce a channel attention mechanism. Instead of dynamic attention, we learn a fixed, parameterized channel interaction matrix $\psi \in \mathbb{R}^{C \times C}$. Each element $\psi_{c, c'}$ represents the learned affinity or score for target channel $c$ attending to source channel $c'$.

During training, we initialize $\psi$ to the identity matrix. This initialization promotes an initial focus on the current channel, reflecting the intuition that a channel is often most relevant to its own past values.

The channel attention weights $W_{\text{chan}} \in \mathbb{R}^{G \times T \times C \times C}$ are computed by applying softmax over the source channel dimension $c'$, after broadcasting $\psi$ and scaling by temperature $\tau$
$$ (W_{\text{chan}})_{g, t, c, c'} = \text{softmax}_{c'}\left( \frac{\psi_{c, :}}{\tau} \right)_{c'}.$$

Note that $\psi_{c, :}$ denotes the $c$-th row of $\psi$, and broadcasting applies $\psi$ uniformly across geo $g$ and time $t$.)

\subsection{Factored Attention}
We combine the temporal and channel attention mechanisms by assuming independence between the temporal and channel selection processes. The temporal map $W_{\text{time}}$ learns \textit{where in time} to attend for each target $(t,c)$, while the channel map $W_{\text{chan}}$ learns \textit{which source channel $c'$} to attend to for each target $(t,c)$.

These marginal attention distributions are multiplied element-wise after appropriate broadcasting to form the final factorized attention weights $W_{\text{fac}} \in \mathbb{R}^{G \times T \times C \times C \times T}$
$$ (W_{\text{fac}})_{g, t, c, c', t'} = (W_{\text{time}})_{g, t, c, t'} \cdot (W_{\text{chan}})_{g, t, c, c'} .$$
This tensor $(W_{\text{fac}})_{g, t, c, c', t'}$ represents the attention weight applied by the target context $(g, t, c)$ to the source value at $(g, t', c')$.

The final output $O \in \mathbb{R}^{G \times T \times C \times D}$ is computed by integrating over the input tensor $X$ using these factorized weights
$$ O_{g, t, c, d} = \sum_{t'=0}^{T-1} \sum_{c' \in \mathcal{C}} (W_{\text{fac}})_{g, t, c, c', t'} \cdot X_{g, t', c', d} .$$

We call this \textit{Factored Attention} because the model learns the channel and time attention components separately before combining them. This factorization reduces the complexity compared to a joint attention mechanism over $T \times C$ positions. If joint attention were used, the attention matrix calculation would scale with $\mathcal{O}((T \cdot C)^2)$, whereas our factored approach scales more favorably with $\mathcal{O}(T^2 + C^2)$ for the core score computations (plus MLP costs).

In some cases, particularly during early model development or for simpler problems, we disable the channel mixing. In this configuration, only the temporal attention $W_{\text{time}}$ is used to weight the values within the same channel
$$ O_{g, t, c, d} = \sum_{t'=0}^{T-1} (W_{\text{time}})_{g, t, c, t'} \cdot X_{g, t', c, d} .$$
This corresponds to the left side of Figure \ref{fig:factored_attn}.

 %end new section

\section{Prediction Heads}
Each input channel in the original data, $X$, has an associated output channel which is generated by a prediction head.  These heads are neural networks that apply the same function to all time-periods and geos.  They are allowed to use a subset of the transformed channels for their estimations.  This allows us to introduce inductive biases and model structures that we believe will aid in model performance and interpretability, while still taking advantage of the complex Transformer transformations from the previous layers.
\subsection{Sales Head}
To predict sales, we start with CoSMo function from \cite{mulc2024compressing} that has already demonstrated to perform well for predicting aggregate U.S. auto sales.  This function utilizes the direction of the Search embedding to yield a probability estimate, which it then scales by the Search volume, $V_{g,t\text{Search}}$.  It takes as input the representation of Search, $X^{(n)}_{g,t, \text{Search}}$, the geo information $g$ (represented as a one-hot vector) as input to a function that outputs a probability estimate\footnote{It also learns a per-geo ``multiplier'' that is uses as a final scaling of the prediction, but it omitted out here for clarity.}.  We want our model to be equivalent to this in the case where the number of Transformer layers is zero and the set of marketing channels is the empty set.  That is we want our model that uses only the Search channel to be

$$ \hat{\text{Sales}}_{g,t} := V_{g,t, \text{Search}} \cdot P(\bar{X}^{(n)}_{g,t, \text{Search}}, g, \theta^{\text{Search}})$$

where
$$\bar{X}^{(n)}_{g,t, \text{Search}}  := \frac{X^{(n)}_{g,t, \text{Search}}}{ ||X^{(n)}_{g,t, \text{Search}}||}$$

and the function $P$ is an MLP Resnet where the final layer has one hidden unit and a sigmoid activation function with parameters $\theta^{\text{Search}}$.  The effect of this roughly mimics a curve of diminishing returns (often modeled by MMMs using a Hill or Weibull function) in that the marginal effectiveness of the impressions on sales saturates.  A key difference, however, is that in MMMs the saturation is with respect to the advertising volume.  In this model, volume scales linearly without saturation, and the output of the saturated ``intent'' function represents the conversion rate of a Search query.

To add multi-channel flexibility in our model, we incorporate the effect of more marketing channels additively\footnote{We experimented with using multiplicative effects via using a log-scale version of the model, but we found the additive models generally performed better and match our expectations that most marketing channels are independent of one another; see the Appendix for more details} as

$$\hat{\text{Sales}}_{g,t} = F(X)_{g,t,\text{sales},0} := \sum_c V_{g, t,c} \cdot P(X^{(n)}_{g,t, \text{search}}, g, \theta^{c}).$$

 In our work, the full set of channels is $\mathcal{C}: =\{$Sales, Search, Search Ads, YouTube Ads$\}$.  We use the same\footnote{A good area for future work will be to explore Hill functions for both Search and non-Search channels, and other well-motivated functions that may take into account the domain knowledge of the channel.} function $P$ for all channels, but with different parameters $\theta^c$.

\subsection{Search Head}
It is generally understood that intermediary organic channels such as Google Search are affected via marketing, but these are often not modeled explicitly.  We aim to incorporate these effects into NNN by generalizing our modeling framework to predict not just sales, but also the Search information.  We do so by predicting the search embedding $X_{g, t+1, \text{Search}}$ at time step $t$ for geo $g$, learning both the magnitude and direction of the search\footnote{This means our model output has values that are representative of the following time-step for the Search channel.}.  This allows us to capture movements in search patterns due to marketing (i.e., movements in the ``baseline'' sales) which are generally treated as exogeneous variables in current MMMs.  Our search head looks like

$$\hat{\text{Search}}_{t, g} = F(X)_{g,t,\text{Search},:} := \text{MLPResnet}(\text{concat}([X^{(n)}_{g, t, c} \mid c \in C'])$$

and takes as a history all previous marketing effort and search data.  Unlike the sales head, we have not given a strong inductive bias to the Search head, because it is less obvious what such a form should entail;  we leave this exploration to future research.  We use an MLP Resnet \cite{resnet, gorishniy2021revisiting} without BatchNorm and Dropout that takes as input the final representation of a subset of channels $C' \subset \mathcal{C}$, which we kept as the YouTube and Search channel in our work, and outputs a $D$-dimensional vector.

\subsection{Other Heads}
The other inputs to our model are marketing channels, which are controlled inputs, whereas the Search and Sales channels are measurements / observations.  Because we treat these as marketing levers and not observations (although it is a viewpoint that Search can affect Search ads, we neglect this in our current models), we don't have a need to predict their values during training.  Out of convenience, to keep the tensor dimensions consistent, we project the intermediate representations of all other channels back to their original size.

\section{Model Training}
We used open-source Bayesian-based MMM methods as a baseline point of comparison, using default out-of-the-box parameters. For the real-world experiments we use priors from a previous training of the model, which replicates a common practice with real-world MMMs.

We train our models with the following loss

$$L(X) := \alpha \cdot L^{\text{sales}}(X, \Theta) + (1 - \alpha) \cdot L^{\text{Search}}(X, \Theta) + \lambda \cdot \sum_{\theta \in \Theta}|\theta|$$

where $\Theta$ is the full set of model parameters, $\lambda$ is the L1 penalty coefficient \cite{lasso}, $\alpha$ is the balancing coefficient between the two losses, and the sales and search losses are further defined as the following mean-squared-errors losses

$$L^{\text{sales}}(X, \Theta) := \frac{1}{G \cdot T} \sum_{g, t} (X_{g, t, \text{sales}, 0} - \hat{\text{Sales}}_{g,t})^2$$

and

$$L^{\text{Search}}(X, \Theta) := \frac{1}{G \cdot (T - 1) \cdot D} \sum_{g, t, d} (X_{g, t+1, \text{search}, d} - \hat{\text{Search}}_{g,t,d})^2$$
where we predict the following time-step's search vector.

We implement our models using JAX, Flax, and Optax \cite{deepmind2020jax}\cite{flax2020github}.  We distribute the model parameters across multiple TPUs (i.e., model sharding).  Most of the models are trained for five-thousand steps with no hyper-parameter tuning on v3 TPUs. For experiments that use hyper-parameter tuning, we perform a grid search using XManager \cite{xmanager}.  All optimizations utilize the full gradient and the default parameters in Adam \cite{kingma2014adam}.  Additionally, we implement gradient clipping \cite{goodfellow} and added a small amount of gradient noise, because the full gradient was used \cite{neelakantan2015adding}.

\section{Sales Attribution}
% We compute the accuracy of the attributed sales estimates from each channel on the simulation data. 

\subsection*{Counterfactual Method}
We compute the attributed sales according to the methods in  \cite{jin2017bayesian} and \cite{meridian_github}, 

$$\text{Attribution}_c = \sum_{g,t} \hat{\text{Sales}}(X)_{g,t} - \hat{\text{Sales}}(\tilde{X}^c)_{g,t}$$

where $X$ represents the original data and media volume and $\tilde{X}^c$ represents the same data except the media from channel $c$ is set to zero.  Because there can be bias errors in predictions, we report the predictions on a mix or percentage basis, defined as $\text{mix}\%_c = \frac{\text{Attribution}_c}{\sum_{c'}\text{Attribution}_c'}$.

\subsection*{Autoregressive Unrolling}
Since our framework also allows us to model the effect on the intermediary organic channels, we can autoregressively unroll predictions out in time.  We start with a prefix $P$ of data and compute the following

$$Y_{P+1} := F(X_{...,:P,...}, \Theta)$$

then for all subsequent $K$ prediction steps we use
$$Y_{P+1+K} := F(\text{concat}([X_{...,:P,...}, Y'_{P+1} ,... Y'_{P+K}], \Theta)$$
where 
$$Y'_t = \text{concat}[Y_{:, t, \text{sales}, :}, Y_{:, t, \text{search}, :}, X _{:, c', \text{search}, :}  | \forall c' \notin \{\text{sales}, \text{search} \}]$$

\begin{figure}
    \centering
    \includegraphics[scale=.36]{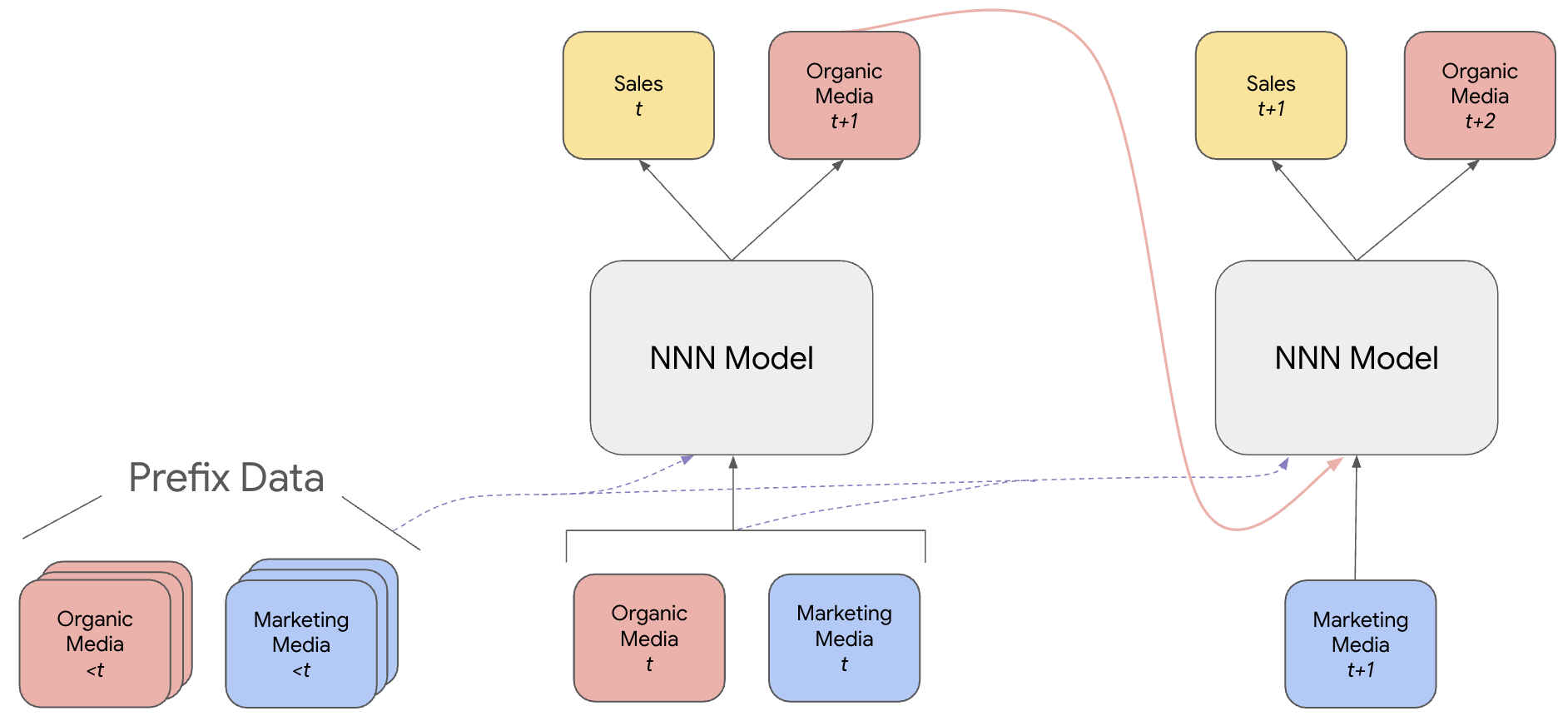}
    \caption{\textbf{Autoregressive decoding}.  Dashed purple lines indicate data that represent the prefix for the current time of decoding.  The model predicts the main target of interest, usually sales, but also the following time-step's organic medic, (e.g., Search), which is then carried over as an input (depicted as the red line) for the next time-step's decoding.  Because our model is not autoregressive with respect to the sales, this value is not carried over.  The input data at time $t$ are then concatenated to the previous prefix to form the new effective prefix for time-step $t+1$.  This process is repeated for the length of the decoding window.  While the sales data is also inputted in the model--albeit immediately masked--and the output out NNN include channels for all marketing channels--albeit not utilized--these are omitted from the figure for clarity.}
    \label{fig:ar_decode}
\end{figure}

This process is pictorially depicted in Figure \ref{fig:ar_decode}.  To attribute sales, we set the given channel after the prefix to a zero volume and follow the same process.  We found that a prefix of 52 weeks is a helpful anchor for the model and is the default value; very short prefix lengths often led to inaccurate attributions, which we believe is due to the data being OOD from the observed training data.  Because of this required prefix, we compute the attribution for multiple 30-week windows and average their results.

\section{Experiments}

When computing sales-prediction error rates, we always roll-up predictions to the national-level to make results easier to interpret.  We evaluate the MAPE and R$^2$ on three datasets: train, validation, and test.  The train dataset is used to fit the parameters, while the other two datasets contain hold-out data.  In all experiments, the test dataset represents data outside the temporal training window (i.e., from dates greater than the last training date), while the validation data is sampled from within the same training window (although this data is not use during model fitting).  Both datasets contain unseen data, and model accuracy measured on either is a proxy for how well the model generalizes, however, generally the test set generalizes with more difficulty, because its data represent samples from a different time period than the data used to fit the model.

\subsection{Simulated Data}
We compare our method, NNN, against a our MMM baseline on both forms of the synthetic dataset.  We measure the predictive error rates on validation and test sets.  Using the models fit only on the train dataset, we calculate the sales attribution of each media channel.  The baseline MMM is fit to the data using the default parameters and priors.  NNN is also fit using its default parameters. We trained two models: the first model is trained using the train data and evaluated on the validation data, while the second model is trained on the train and validation data and evaluated on the test data.  Thus we show a pair of values that corresponding to training metrics for MMM.  We compute the mix\% for each channel and each model.  For NNN, we train two versions of the model: one using only the Sales as the target and another using both Search and sales as the target.  Both models are trained without channel mixing for 5k steps.  For the Search and sales model, we train for an additional 5k steps using balancing coefficient of .99, and we report the mix\% using both the traditional method and the autoregressive (AR) unrolling method. 

\subsubsection*{Predictive Errors}
Tables \ref{tab:error_low} and \ref{tab:error_high} show the predictive errors for models trained on the Low Variance and High Variances datasets, respectively. It should be noted that MMMs are not usually evaluated primarily in terms of their predictive error since they are not predictive models; instead MMMs are explanatory models and/or designed for causal inference. In the NNN framework that we propose in this paper, however, the NNNs can be treated as predictive models so we adopt that comparison here.

For both datasets, NNN outperforms our baseline MMM on both holdout sets in terms of R$^2$ and MAPE.  For the Low Variance dataset, both models achieve good MAPE metrics on all data splits, however NNN does a much better job at explaining the variance of the data, as measured by R$^2$.  Interestingly, NNN achieves MAPE values that may be approaching values generated by the synthetic noise, without overfitting to the train dataset. For the High Variance dataset, NNN outperforms MMM in both MAPE and R$^2$ as well.  Most of the variance in the sales can be measured using a subset of the Search terms (i.e., the high-impact terms), which NNN appears to correctly parse using the embeddings, while MMM, which only uses the Search volume, is unable to differentiate between the high-impact and low-impact terms. 

\subsubsection*{Sales Attribution Estimates}
Tables \ref{tab:roi_combined_gt_last} show the estimated sales attributions for models trained on the Low Variance and High Variances datasets, respectively.  We show two versions of the ground truth: Direct and Total.  The direct version is the mix that appears to be causing the sales; YouTube impressions that are driving Search (the line from YouTube to Search in Figure \ref{fig:sim_dag}) which drive sales get counted as Search sales not YouTube sales, and only the sales directly attributed to YouTube (the blue line Figure \ref{fig:sim_dag}) are counted towards YouTube.  The total version counts sales that drive Search which then drive sales as YouTube sales (indirect sales), in addition to the direct YouTube sales.

For both datasets, the baseline MMM provides decent estimates across all channels. The NNN is closer to the ground truth for YouTube and for Search, while the MMM is closer to ground truth for Search Ads. Both models under-predict the attribution of Search Ads, and both models are closer to the Direct values than the Total values for the Search and YouTube channels, as expected. When utilizing AR unrolling, we see that NNN properly moves sales from Search to YouTube, however it slightly underestimates the YouTube contribution (39.65\%) for the High Variance Dataset, and it slightly overestimates for the Low Variance Dataset (55.10\%).  Interestingly, it also adjusts the Search Ads mix estimates. We leave further investigating to future work\footnote{Particularly in the area of more structural Search heads and a robust method for determining the balancing coefficient.}, although we are excited about the promise of this direction.

\begin{table}[htbp] 
\centering
\begin{tabular}{l rr rr rr} % l=left align method, r=right align numbers
\toprule % Top rule from booktabs
          & \multicolumn{2}{c}{Train} & \multicolumn{2}{c}{Validation} & \multicolumn{2}{c}{Test} \\ % Centered multi-col headers
\cmidrule(lr){2-3} \cmidrule(lr){4-5} \cmidrule(lr){6-7} % Rules under multi-col headers
Method    & MAPE      & R$^2$         & MAPE       & R$^2$         & MAPE      & R$^2$       \\ % Aligned sub-headers
\midrule % Mid rule from booktabs
MMM       & [3\% -- 3\%] & [0.28 -- 0.76] & 3\%        & 0.55          & 3\%       & 0.44        \\
NNN       & \textbf{2\%} & \textbf{0.95}  & \textbf{1\%} & \textbf{0.94} & \textbf{1\%} & \textbf{0.95} \\ % Use \sim for approximation
\bottomrule % Bottom rule from booktabs
\end{tabular}
\caption{Model error rates (MAPE and R$^2$) for MMM and NNN on the simulated Low Variance dataset. Bolded values denote better performance.}
\label{tab:error_low} 
\end{table}

\begin{table}[htbp] 
\centering
\begin{tabular}{l rr rr rr} % l=left align method, r=right align numbers
\toprule % Top rule from booktabs
          & \multicolumn{2}{c}{Train} & \multicolumn{2}{c}{Validation} & \multicolumn{2}{c}{Test} \\ % Centered multi-col headers
\cmidrule(lr){2-3} \cmidrule(lr){4-5} \cmidrule(lr){6-7} % Rules under multi-col headers
Method    & MAPE       & R$^2$        & MAPE        & R$^2$         & MAPE      & R$^2$       \\ 
\midrule % Mid rule from booktabs
MMM       & [15\% -- 15\%] & [0.23 -- 0.36] & 15\%        & 0.26          & 15\%      & 0.21        \\
NNN       & \textbf{3.5\%} & \textbf{0.96}  & \textbf{2.6\%} & \textbf{0.97} & \textbf{3.2\%} & \textbf{0.96} \\ 
\bottomrule % Bottom rule from booktabs
\end{tabular}
\caption{Model error rates (MAPE and R$^2$) for MMM and NNN on the simulated High Variance dataset. Bolded values denote better performance.}
\label{tab:error_high} 
\end{table}

% (Injecting a lot of variance just through the embeddings, which R$^2$ is a measure of, and MMM is not able to capture).

\begin{figure}
    \centering
        \centering
        \includegraphics[width=.8 \textwidth]{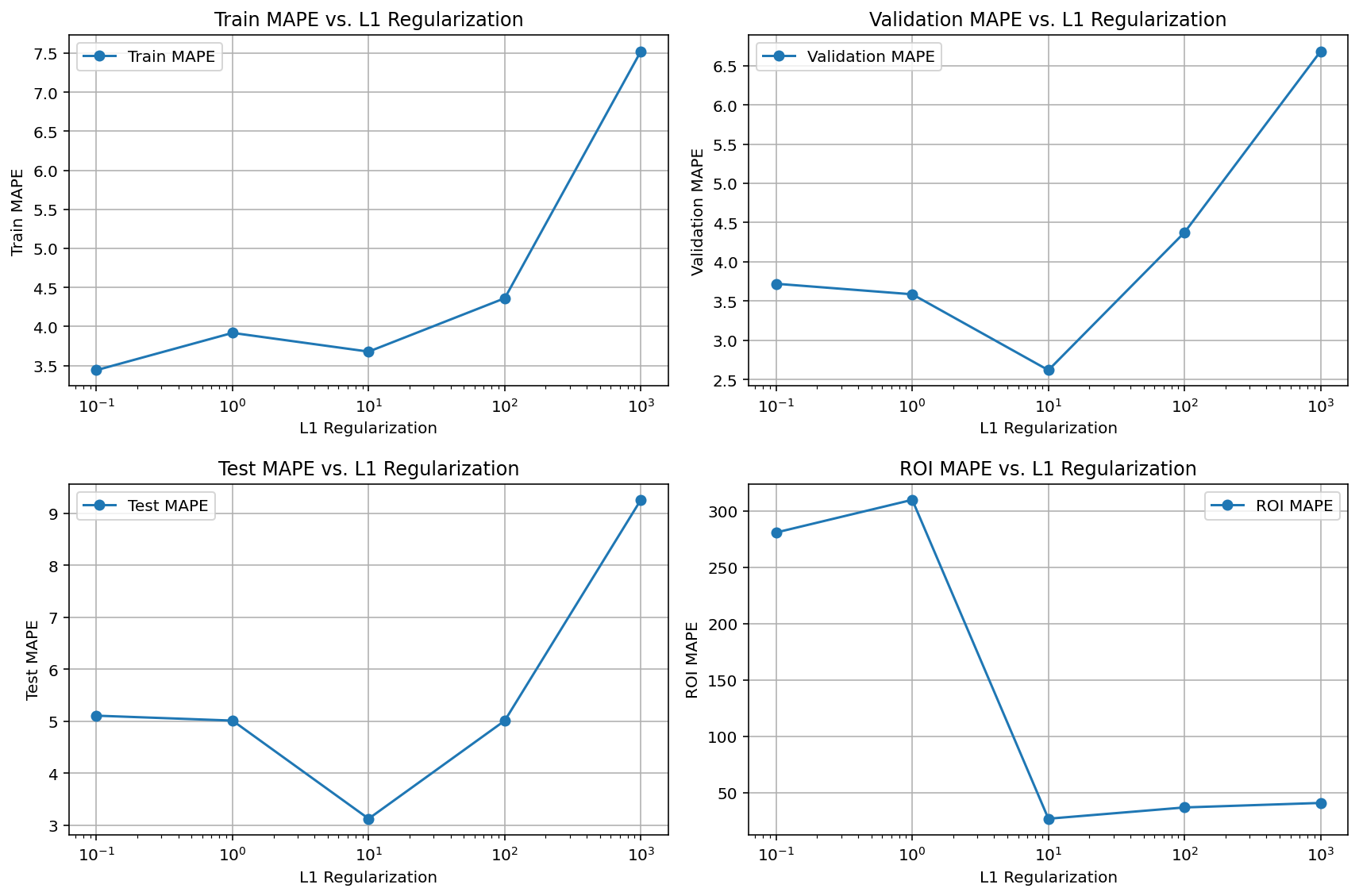} 
    \caption{Relationships between L1 regularization, attribution estimates, and out-of-distribution error rates.  Picking models with low test and validation error rates--largely determined by the L1 term--yields good attribution estimates.  When the models exhibit overfitting (when regularization is too low), their attribution estimates tend to suffer greatly.  Experiments are from the High Variance Dataset.  The Low Variance experiments are shown in Figure \ref{fig:low_var_l1} the Appendix and follow the same pattern.}
    \label{fig:l1_vs_roi}
\end{figure}

\begin{table}[htbp]
\centering
\begin{tabular}{l | ccc | ccc}
\toprule
 & \multicolumn{3}{c|}{High Variance Dataset} & \multicolumn{3}{c}{Low Variance Dataset} \\
\cmidrule(lr){2-4} \cmidrule(lr){5-7} %
Method & Search & Search Ads & YouTube & Search & Search Ads & YouTube \\
\midrule\midrule
% --- Model Estimates ---
MMM & 77.62\% & \textbf{7.03\%} & 15.35\% & 74.31\% & \textbf{6.19\%} & 19.50\% \\
NNN & 68.71\% & 4.19\% & 27.10\% & 71.96\% & 3.93\% & 24.10\% \\
\begin{tabular}[c]{@{}l@{}}NNN (Search + \\ Sales training)\end{tabular} & 65.62\% & 3.81\% & 30.57\% & 71.78\% & 3.24\% & 24.91\% \\
\midrule 
\begin{tabular}[c]{@{}l@{}}NNN\\ (AR Unroll)\end{tabular} & \textbf{29.88\%} & 30.46\% & \textbf{39.65\%} & \textbf{22.72\%} & 22.18\% & \textbf{55.10\%} \\
\midrule % Rule separating AR Unroll from Ground Truth
% --- Ground Truth (Moved to Bottom) ---
\begin{tabular}[c]{@{}l@{}}Direct\\ Ground Truth\end{tabular} & 59.14\% & 10.03\% & 30.83\% & 59.14\% & 10.03\% & 30.83\% \\
\begin{tabular}[c]{@{}l@{}}Total\\ Ground Truth\end{tabular} & 43.68\% & 10.03\% & 46.29\% & 43.68\% & 10.03\% & 46.29\% \\
\bottomrule % Final rule
\end{tabular}
% \label{tab:roi_combined_gt_last}
\caption{Sales Attribution Mix Estimates Comparison for High and Low Variance Datasets. Bolded values denote better performance (e.g. closer to the Total Ground Truth).}
\label{tab:roi_combined_gt_last}
\end{table}

\subsubsection*{The Role of L1 Regularization in Model Selection} 

To understand the impact of regularization on NNN performance and attribution, we conducted a grid search over the L1 penalty coefficient ($\lambda$), evaluating both predictive errors (MAPE, R$^2$) across train, validation, and test splits, as well as the accuracy of sales attribution estimates against ground truth in our simulations. The relationship between regularization strength, predictive error, and attribution error is illustrated in Figure \ref{fig:l1_vs_roi}. 

Our central observation is a strong relationship between out-of-distribution (OOD) predictive accuracy and attribution accuracy. We find that as L1 regularization is relaxed (i.e., $\lambda$ decreases), both predictive error rates and attribution error generally decrease in tandem up to a critical point. Beyond this point, typically characterized by the onset of overfitting where OOD predictive errors (validation and especially test set errors) begin to rise, attribution accuracy sharply degrades. Concurrently, as regularization decreases and OOD error increases, model sparsity also significantly decreases, indicating the model is using more non-zero parameters (see Figure \ref{fig:l1-reg} in Appendix). 

This finding suggests that minimizing predictive error on appropriately chosen OOD datasets is a promising strategy for identifying NNN models that also yield accurate channel attributions. This perspective contrasts somewhat with the way that MMMs are generally built \cite{chan2017challenges}, because MMMs are generally explanatory or causal models rather than predictive models. Thus, within that framework, MMM authors often caution against relying solely on statistical fit metrics for model selection. We hypothesize that NNN's success with this OOD-driven approach stems from its evaluation across multiple hold-out sets, particularly the use of a test dataset drawn from a \textit{temporally distinct} period from the training data. This provides a more challenging and informative signal for generalization compared to evaluations restricted to in-sample data or temporally similar validation sets. Furthermore, the sparsity induced by L1 regularization appears particularly beneficial for managing the large parameter spaces inherent in neural network architectures, especially within the often data-constrained context of MMM.

From a practical standpoint, this relationship between OOD performance and attribution accuracy potentially streamlines the model selection workflow. Instead of potentially lengthy, iterative cycles of selecting priors, retraining, and evaluating based on parameter plausibility and fit, our approach centers on tuning a single key hyperparameter ($\lambda$) via a readily parallelizable grid search. Models can be primarily selected based on achieving optimal predictive performance on OOD validation/test sets. While this data-driven selection forms the core, we recommend that chosen candidate models still undergo verification through attribution analysis and the simulation-based probing discussed previously and below, ensuring the selected model behaves reasonably beyond aggregate predictive metrics. As we will discuss later in this paper, future validation and research will be required in order to assess the robustness and quantify the uncertainty distribution of measurements made under the NNN framework.

\subsubsection{Model Application beyond the MMM framework: Simulation and Analysis}
Once trained, the NNN model serves not only as a predictive tool but also as a flexible engine for simulation and analysis. By systematically manipulating specific components of the input tensor $X$—representing counterfactual scenarios (e.g., changing marketing volume/content) or isolating specific points of interest (e.g., individual keyword embeddings)—and observing the resulting changes in the model's outputs (e.g., predicted sale), we can probe the learned relationships and estimate the impact of various interventions or input characteristics. This inference-based analysis allows for explorations beyond standard forecasting or aggregate channel attribution, providing deeper insights into the system dynamics captured by the model. The following subsections detail two primary applications of this capability demonstrated in our work: simulating the effects of marketing pauses and analyzing the inferred value of specific creative or keyword embeddings.

\textbf{Predictions}
As \cite{mulc2024compressing} demonstrated, the representation of the weighted search embedding and the search volume in these models can be incredibly effective prediction tools. We expect the same results would hold here, with the addition of including the effects of media on sales in addition to organic search.

\textbf{Marketing Pause}
We simulate the effect of a YouTube pause by defining a twenty-week period to set the YouTube marketing volume to zero.  We then compute model inference two ways: using autoregressive unrolling and using the standard inference with the observed values of Search.  This simulation is shown in Figure \ref{fig:yt_pause}.  The estimated sales for both pause experiments are smaller than the estimated sales using the observed marketing and search data (i.e., no pause), however the effect of the pause is much more pronounced when accounting for the effect YouTube had on Search, and the pause effect lasts much longer.

\textbf{Creative and Keyword Analysis via Model Inference} Beyond forecasting, our trained model enables insightful analysis of creative elements or keywords without requiring separate A/B tests or experiments. Once trained, our model implicitly learns the relative value associated with different input features represented as embeddings. We can leverage this by performing inference on specific embeddings corresponding to distinct creatives (like videos for YouTube; i.e., content) or keywords (for Search or Search Ads channels; i.e., context), similar to the approach in \cite{mulc2024compressing}. By analyzing the model's output score associated with these specific embeddings, we can infer their relative effectiveness.

To ensure the statistics of the embeddings fed during inference align closely with those observed during training, we introduce a specific input construction process. For analyzing an embedding $v \in \mathbb{R}^D$ associated with a target channel $c_{\text{target}}$ (e.g., Search), we first compute a global anchor embedding $A \in \mathbb{R}^D$. This anchor represents the average state of the target channel across all geos $G$ and time $T$:
$$ A = \frac{1}{G \cdot T} \sum_{g,t} X_{g, t, c_{\text{target}}, :} $$
We also compute per-geo contextual embeddings $\mu^{k}_g \in \mathbb{R}^D$ for other relevant channels $k \neq c_{\text{target}}$ by averaging over time for each geo $g$:
$$ \mu^{k}_g = \frac{1}{T} \sum_{t} X_{g, t, k, :} $$
These contextual embeddings provide the model with the typical background state of other channels when evaluating the target embedding $v$.

The core input feature for the target channel $c_{\text{target}}$ and geo $g$ is then constructed by scaling the specific embedding $v$ and centering it around the anchor $A$:
$$ v'_{g} = \alpha v + A $$
where $\alpha$ is a scaling factor typically on the order of a standard deviation of the norm of the target embeddings seen during training. Note that $v$ and $A$ are broadcast to match across the geo dimension $G$ before the element-wise addition.

This anchored embedding $v'_{g}$ is then placed into the appropriate channel slot $c_{\text{target}}$ within a larger input tensor $X_{\text{in}} \in \mathbb{R}^{G \times 1 \times C \times D}$ used for model inference. The slots for the contextual channels $k$ are filled with their respective per-geo means $\mu^{k}_g$. This construction effectively evaluates the impact of the specific (scaled) embedding $v$, conditioned on the average behavior of other channels $\mu^k_g$.

The final score $S(v)$ associated with embedding $v$ is obtained by applying the model to the constructed input $X_{\text{in}}(v)$ and extracting and aggregating a relevant part of the model's output $ F(X_{\text{in}}(v), \Theta)$, such as sales $\sum_g F(...)_{g,0,\text{sales},0}$

\begin{figure}
    \centering
    \includegraphics[scale=.38]{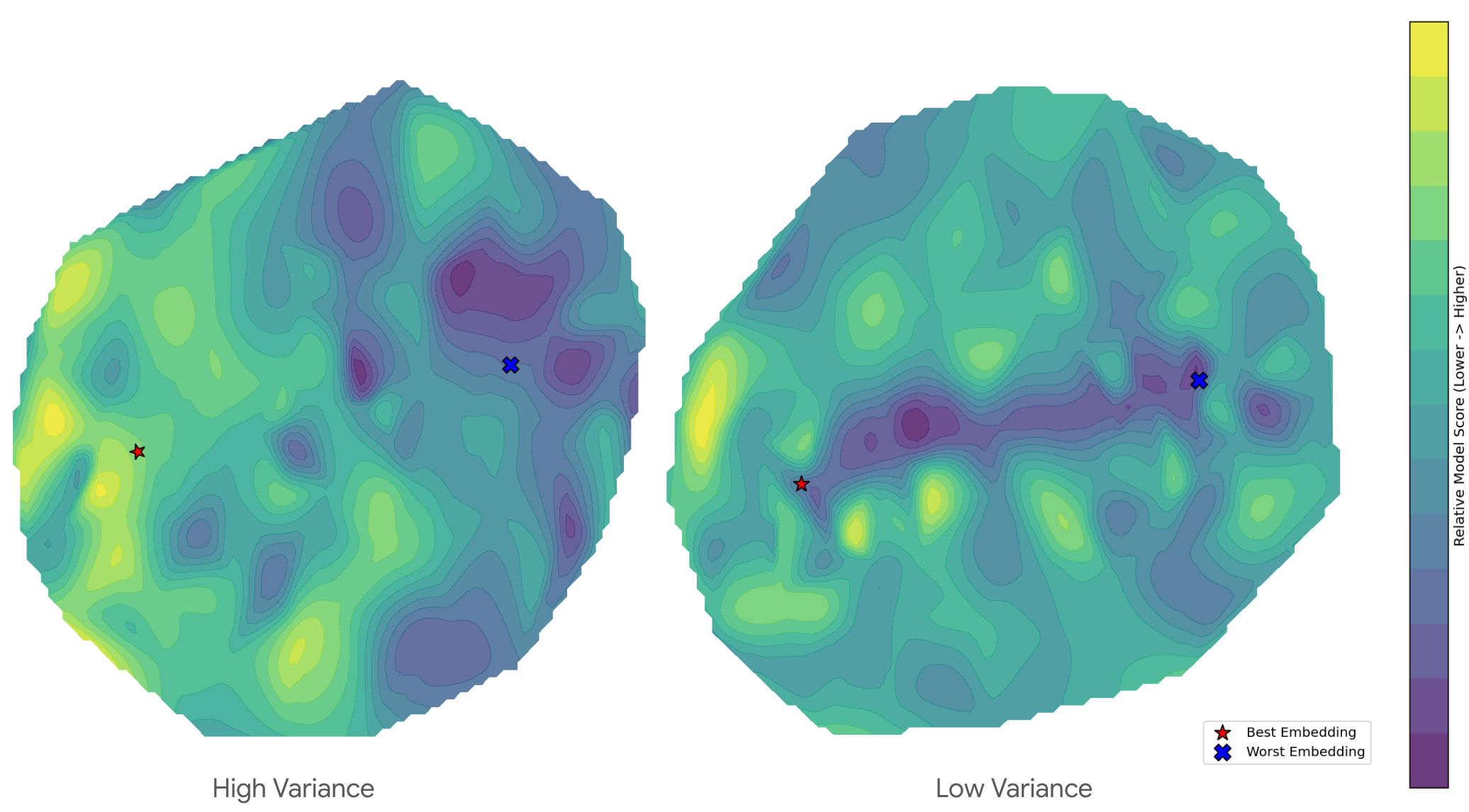}
    \caption{Model score landscape for the Search channel, projected into 2D using t-SNE for the High Variance Dataset (Left) and the Low Variance Dataset (Right). The heatmap indicates the predicted sales impact, with known high-impact ('Best') and low-impact ('Worst') query embeddings overlaid.}
    \label{fig:high_var_landscape}
\end{figure}

Figure \ref{fig:high_var_landscape} demonstrates this analysis for the Search channel. We generated a landscape of scores $S(v)$ for various embeddings $v$ sampled around known `best` and `worst` performing query embeddings (identified by their correlation with sales increases, known \textit{a priori} since the dataset was manufactured). The embeddings were projected to 2D using t-SNE, and the scores are visualized as a heatmap. The plot confirms that the model assigns higher scores to embeddings near the `best` query vector and lower scores near the `worst` vector, demonstrating the model's ability to discern embedding quality.

However, we note that the ability to glean these insights likely depends on the amount of variation present in the training data. We contrast the results in Figure \ref{fig:high_var_landscape} for the High Variance Dataset with those from the same model trained on the Low Variance Dataset. As observed in the Low Variance landscape, the model assigns only marginally higher scores to the 'best' query vector compared to the 'worst', appearing largely unable to discern query quality. We posit that because the low variance dataset lacks sufficient keyword variation, the model cannot effectively learn the correlation between query content and sales.

This analysis framework can be applied to any channel for which specific embeddings (creatives, keywords, etc.) are available. When using such channels, it provides valuable insights into creative or keyword effectiveness in parallel to media measurement for these channels. However, NNN does require these embeddings in order to generate these additional insights and use the content or the context of the ads for its modeling (and for many traditional media channels like TV these will not be readily available). We have not yet done comprehensive research on the effectiveness of this framework with purely scalar channels, especially at the scale of dozens of channels which is common for MMMs.

\subsection{Real World Marketing Data}
First, we compare our baseline MMM against our NNN method on a real world dataset. Both models were trained at a regional level (4 regions were used, as defined by U.S. Census), and the data was at a weekly granularity. 

Previously, an MMM was fit for our end user, and we use the formerly informed priors to fit this version of the model.  We used standard fitting methods used in practice.  A time-variant effect was applied to this MMM (at the end of each month we insert one coefficient). In order to give a fair comparison on the test set, which appears after all training and validation points, we allowed the MMM to fit two additional points to the baseline during this period.\footnote{This effectively allowed the MMM to observe the holdout set during fitting, but it was the only way we could see giving a somewhat comparable estimation on data for test the timeframe.}.  We used 7 chains with 1000 burn-ins, and all the channels converged during optimization.  We only tuned the standard deviations of the priors while looking at model diagnostics and convergence (i.e., if the baseline had a negative contribution, we retrained).  Similar to the simulated datasets, we followed a two-model training procedure for validation and test evaluations.

NNN was fit using the default hyper parameters.  We include a baseline model that uses only Search and sales data.  The full models uses sales, Search, and two top attribution channels.  These additional two channels are included only using their impressions as scalar quantities, and no qualitative embedding information is used.  The results are shown in Table \ref{tab:mmm_vs_nnn}.

We notice that the baseline MMM does a good job at predicting out of sample data during the training period (validation MAPE 20\% and R$^2$ .77), however it suffers during the test period (189\% MAPE, .90 R$^2$).  The NNN Baseline generalizes well in both the validation data (5.3\% MAPE, .99 R$^2$) and the test data (16.2\% MAPE, .92 R$^2$), achieving better errors than MMM.  When the additional channels are included, the validation and test errors are further reduced, demonstrating that NNN can make effective use of the additional channels using the same format of data traditionally available in MMM.

\begin{table}[htbp] % Use placement specifier like [htbp]
\centering
% Using 'l' for the first column (Model) and 'r' for all numeric/data columns
\begin{tabular}{l rr rr rr}
\toprule % Top rule from booktabs
             & \multicolumn{2}{c}{Train} & \multicolumn{2}{c}{Val} & \multicolumn{2}{c}{Test} \\ % Centered multi-col headers
\cmidrule(lr){2-3} \cmidrule(lr){4-5} \cmidrule(lr){6-7} % Rules under multi-col headers
Model        & MAPE      & R$^2$        & MAPE          & R$^2$      & MAPE           & R$^2$      \\ % Right-aligned sub-headers (unit is now in data)
\midrule % Mid rule from booktabs
MMM          & 18.0\%    & 0.87         & 20.0\%        & 0.77       & 189.0\%        & -90        \\ % Added % to MAPE data
NNN Baseline & \textbf{5.3\%} & 0.99    & 16.2\%        & 0.92       & 22.5\%         & 0.95       \\ % Added % to MAPE data
NNN          & 8.9\%     & \textbf{0.99}& \textbf{13.0\%} & \textbf{0.93} & \textbf{10.0\%} & \textbf{0.98} \\ % Added % to MAPE data
\bottomrule % Bottom rule from booktabs
\end{tabular}
\caption{MMM vs NNN on a real-world dataset. The baseline uses only organic Google Search queries as sales input, while the full model uses two additional paid media channels without qualitative embedding information.} % Removed sentence about units, as % is explicit now
\label{tab:mmm_vs_nnn} % Changed label slightly
\end{table}

\subsubsection{Additional Experiments}

We further test our method on weekly, but DMA-version (designated marketing area) of the Real World dataset using the embedded versions of Search Ads and YouTube.  For all experiments, we run a grid search on the L1 values and pick the best model according to the validation MAPE.  The results are shown in Table \ref{tab:dma_performance}.   We see that as we add more channels to the model, performance generally improves without overfitting.  The best overall improvements (measured by average rank across all four metrics) on the validation set come from adding the YT channel, while for the test set, the most improvements come from adding the Search ads channel, although using all channels performs very similarly.  Interestingly, adding the Search training objective hurts the baseline model (Search only), but generally helps or does no harm to the other models.  We believe for the baseline model, because the only signal comes from Search, the model is essentially over regularized, where as for the other models, the parameters related to the marketing channels can be used to lower the Search error without harming sales component of the model.

\begin{table}[h!]
    \centering
    \begin{tabular}{lcc|cc|ccc|c}
        \toprule
        & Val MAPE & Val R$^2$ & Test MAPE & Test R$^2$ & Search & Search Ads & YouTube & Search Training \\
        \hline
        & 17.97 & 0.8256 & 33.35 & 0.8490 & \checkmark & & &  \\
        & 15.44 & 0.9349 & 72.35 & 0.7142 & \checkmark & & & \checkmark \\
        & \textbf{13.15} & 0.9500 & 61.86 & 0.7701 & \checkmark & & \checkmark & \\
        & 13.84 & \textbf{0.9717} & 33.36 & 0.8908 & \checkmark &  & \checkmark & \checkmark \\
        & 15.50 & 0.9232 & 49.37 & 0.8257 & \checkmark & \checkmark & & \\
        & 20.62 & 0.8945 & \textbf{25.89} & \textbf{0.9088} & \checkmark & \checkmark & & \checkmark \\
        & 13.87 & 0.9072 & 40.90 & 0.9026 & \checkmark & \checkmark & \checkmark &  \\
        & 15.96 & 0.9621 & 42.77 & 0.8675 & \checkmark & \checkmark & \checkmark & \checkmark \\
        \bottomrule
    \end{tabular}
    \caption{DMA-level Model Performance Metrics}
    \label{tab:dma_performance}
\end{table}

\begin{figure}
    \centering
    \includegraphics[scale=.28]{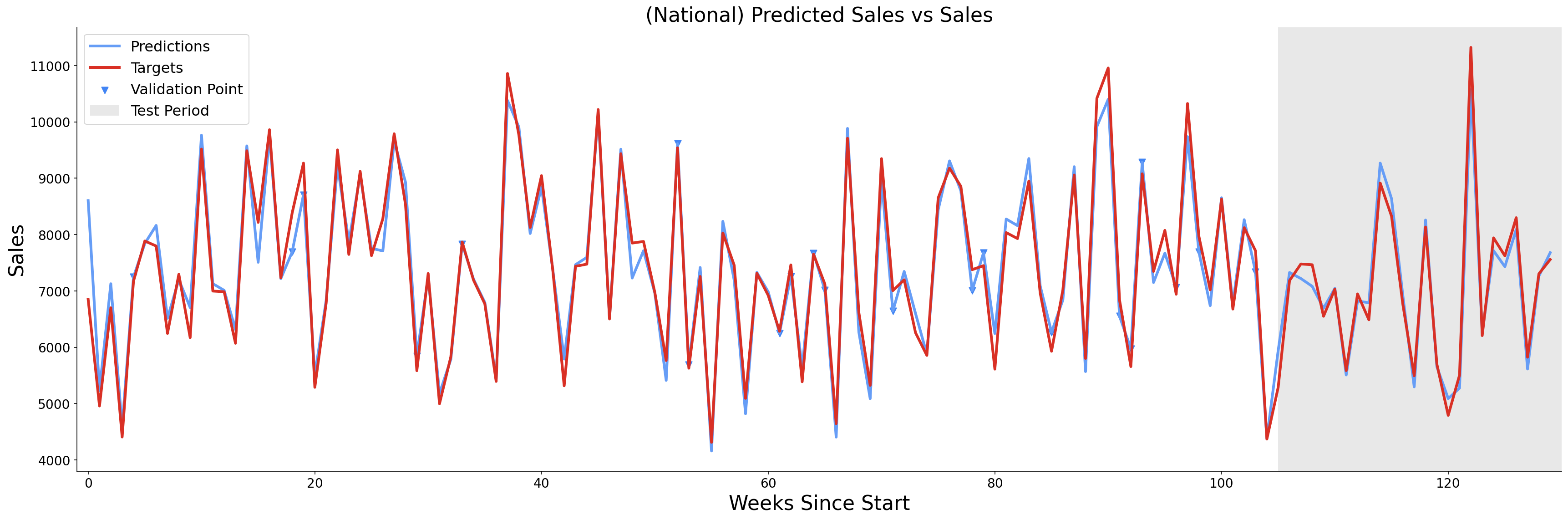}
    \caption{Actual sales and predicted sales (rolled up to National level) over time using only for the High Variance dataset.  Model trained using all three channels using only sales as a target.}
    \label{fig:predicted_Sales}
\end{figure}

\begin{figure}
    \centering
    \includegraphics[scale=.28]{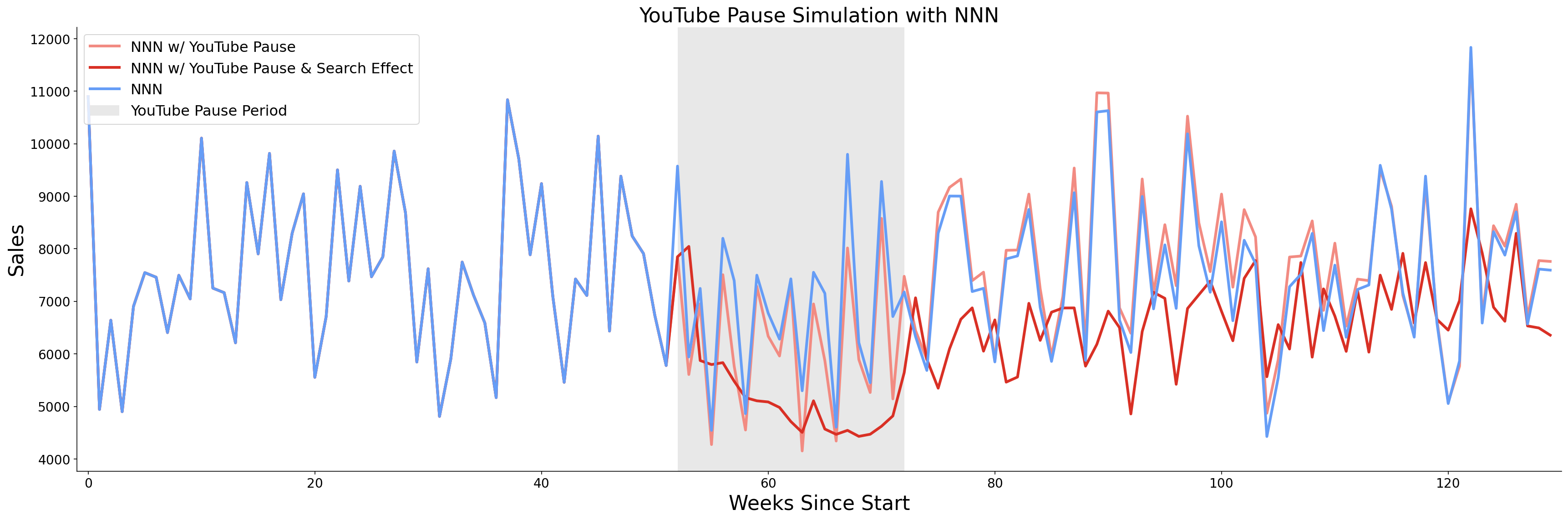}
    \caption{Simulated effect of a 20-week YouTube pause on the High Variance Dataset.  Plot was produced using a model with four Transformer layers that was trained on both Search and sales.}
    \label{fig:yt_pause}
\end{figure}

\section{Discussion: NNN in the Context of MMM}

The introduction of NNN presents a significant departure from general MMM methodologies. By leveraging the representational power of neural networks, Transformer architectures, and rich embeddings of marketing channels alongside organic signals like Google Search, NNN offers a novel framework for modeling marketing effectiveness. This section compares NNN to standard MMM approaches across several key dimensions, highlighting differences in methodology and capability.

\textbf{Modeling Foundations.}
NNN sits outside the established statistical frameworks used for MMMs. As we mentioned above, while some of its functionality overlaps with the types of media measurement that MMMs provide, it does not necessarily replace MMM in real-world contexts. As we discuss below, NNNs require significant additional research and validation. 

MMMs typically employ established statistical models, often based on linear regression, incorporating transformations like adstock and Hill functions to capture non-linearities and carryover effects. NNN, in contrast, NNN utilizes the inherent non-linear function approximation capabilities of deep neural networks and the sequence modeling power of Transformers and self-attention. This allows NNN to potentially capture more complex interactions and temporal dependencies directly from the data, although again more research is required here. 

Furthermore, MMMs primarily rely on scalar time-series representations of marketing inputs (e.g., weekly spend or impressions). NNN uses these data as well, but supplements them with rich, high-dimensional embeddings. These embeddings integrate both quantitative metrics (like impressions, encoded in the embedding's magnitude) and qualitative aspects (like ad creative content or search query semantics, encoded in the embedding's direction). With this semantic information, the hope is that NNN can distinguish between, for example, a million impressions of a brilliant ad versus a million impressions of a dud. The performance difference observed on the High Variance simulated dataset, where the semantic nuance was deliberately included, underscores the value of this approach.

\textbf{Incorporating Prior Beliefs: Distributions vs. Architecture vs. Optimization.} A fundamental difference lies in how prior knowledge or beliefs are incorporated. Bayesian MMMs directly embrace the concept through explicit prior distributions assigned to model parameters. The modeler specifies these distributions, drawing perhaps on previous experiments, established benchmarks, or, in the absence of strong information, resorting to non-informative priors. This allows for granular control and the injection of domain expertise directly into parameter estimation.

NNN, on the other hand, largely eschews this kind of parameter-level prior specification. While its network weights are initialized before training, the methods typically used—standard schemes like Glorot and He initialization\cite{glorot2010understanding, he2015delving}—are chosen primarily to ensure stable gradient flow and effective optimization, rather than to encode any specific belief about the final value of a given weight parameter\footnote{Although there are cases where special care is taken, such as the identity matrix initialization for the channels matrix in our attention mechanism.}. In general, there are too many parameters, interacting in too complex a way, for individual NNN weights to be easily interpretable or specifiable in a Bayesian sense. Instead, in NNN, prior beliefs are encoded more implicitly, baked directly into the architecture and computational structure of the model itself. For example, when designing our Transformer's attention mechanism, the use of a causal mask rigidly enforces the prior belief that future events cannot influence past predictions. Similarly, the deliberate removal of layer normalization components within the Transformer stems from the prior understanding that the magnitude of our embedding vectors carries crucial information about marketing volume, information that normalization might otherwise discard. These architectural choices act as strong structural priors, shaping the function space the model explores, even if individual parameters aren't given explicit Bayesian priors.

Beyond these architectural choices, the optimization process itself introduces another form of implicit prior via regularization. Our NNN framework employs an L1 penalty added to the loss function during training. This L1 term actively encourages sparsity by pushing many model parameters towards zero. This effectively embeds a belief in sparsity within the learning process itself – an assumption that, within the potentially vast parameter space of the neural network, many connections or features are ultimately unnecessary for explaining the data. However, this sparsity-inducing prior is typically chosen more for its beneficial regularization effects—combating overfitting in high dimensions and aiding model selection via OOD performance-than for encoding specific, quantifiable beliefs about the likely value of any individual parameter, unlike the explicit distributional priors of Bayesian models.

\textbf{Model Uncertainty.}
Bayesian MMMs offer a significant advantage in inherently quantifying uncertainty through posterior distributions for parameters and predictions.  By using priors, and MCMC sampling, they provide not just point estimates, but posterior distributions for model parameters and predictions.  This furnishes Bayesian MMMs with a statistically grounded measure of confidence. In constrast NNN is trained by providing point estimates of sales and media channel values, then using gradient-based optimization moves the point estimate of the parameters. This provides neither uncertainty estimates of the model parameters nor the predictions.  Because the NNN model is a neural network, providing distributions over the parameters won't make much practical difference over the current point estimates: these parameters are usually not inspected.  However, providing uncertainty distributions for the predictions is useful, and we look to techniques like Monte Carlo Dropout as a viable path for incorporating uncertainty estimation in future work.

\textbf{Capturing Temporal Dynamics: Seasonality, Baselines, and Long-Term Effects.}
Handling temporal patterns is critical in marketing modeling. Many MMMs address seasonality using explicit indicator variables or spline knots, while carryover is modeled with parametric adstock functions, which typically struggle to capture effects beyond a few months. Baseline demand is often represented by simple trend components.

NNN takes a different approach by leveraging rich, dynamic signals and its sequence modeling architecture. We rely on organic signals, particularly Google Search embeddings, to implicitly capture complex baseline fluctuations, including seasonality. As prior work suggests, Search behavior naturally reflects weekly, seasonal, and holiday-related patterns. NNN's neural networks can parse this relevant information directly from the Search embeddings, reducing the need for extensive seasonal feature engineering. Moreover, by explicitly modeling the influence of marketing on intermediary signals like Search and employing a Transformer architecture with its inherent attention mechanism over long sequences, NNN has the potential to capture marketing effects unfolding over much longer time horizons than traditional adstock allows. Our simulation experiments, particularly the YouTube pause analysis (Figure \ref{fig:yt_pause}), provide evidence for this capability, although further research and validation is needed. 

\textbf{Attribution and Interpretation.}
Both NNN and MMMs generally resort to counterfactual simulations—running the model with and without a specific channel's activity—to estimate its contribution. However MMMs also can provide direct interpretability through parameter estimates (e.g., coefficients, decay rates, saturation parameters) within well-defined functional forms.

NNN, like many deep learning models, faces challenges in direct parameter interpretability. However, we believe its strength lies in modeling the dynamic interplay between selected channels and its ability to perform analyses like the creative/keyword analysis. By probing the model's response to specific, meaningful inputs (like different ad embeddings), we can gain valuable insights into the learned relationships and drivers of effectiveness, offering a different but complementary form of interpretability. Furthermore, modeling intermediary channels like Search allows for potentially more nuanced attribution, distinguishing between direct effects on sales and indirect effects mediated through organic demand shifts.

\textbf{Model Selection.}
Finally, the process of building and selecting models also differs. MMMs often involve a significant degree of human judgment, balancing statistical fit against the plausibility of parameters (do they fall within expected ranges based on prior knowledge or beliefs?) and the overall interpretability of the resulting story. While objective metrics are used, they often serve as guardrails rather than the sole objective function, partly due to concerns about overfitting and the potential for multiple different models to fit historical data equally well.

NNN, a machine learning-forward approach, leans more heavily on objective, out-of-distribution (OOD) predictive performance as its North Star. Metrics like MAPE and R$^2$ calculated on held-out validation set and, crucially, a temporally distinct test set are the primary criteria for selecting hyperparameters, particularly the L1 regularization strength which governs model sparsity. The rationale, supported by our experiments, is that models generalizing well to unseen, OOD data are more likely to have captured robust underlying patterns and thus yield more reliable attribution estimates, even if their internal parameters are complex or non-intuitive. This data-driven selection process, which can involve parallel model fitting, can be significantly faster than iterative prior adjustment.

However, given the novelty of the approach, the limited set of examples examined so far in this paper, and the inherent risks of complex models, we strongly recommend supplementing these OOD performance checks with business judgment and context. This can be introduced through the kinds of simulation-based probing (attribution, creative analysis) discussed earlier as a crucial vetting step.

\textbf{Summary.}
As it covers a slightly different set of potential use cases, NNN introduces a different set of trade-offs compared to MMM. It seems to excel in leveraging rich embedding representations, modeling complex non-linearities and long-term temporal dynamics via its neural and Transformer components, and utilizing predictive performance on holdout data for model selection. Current MMM approaches, particularly Bayesian MMMs, hold advantages in direct parameter interpretability and built-in uncertainty quantification. NNN represents a promising research direction, especially suited for scenarios rich in contextual data where capturing complex interactions and long-range effects is paramount, though further validation and development is needed.

\subsection{Future Directions}

The promising results of this initial exploration of NNN suggest several important avenues for future research and development aimed at validating the methodology and extending its capabilities and robustness:

\textbf{Additional Validation Required} While simulations and retrospective analyses are valuable, the practical utility of NNN must be confirmed through rigorous, live marketing experiments. Comparing NNN’s performance, attribution results, and potential optimization recommendations directly against established MMMs in real-world settings combined with live experimentation is essential for comprehensive validation and identifying areas requiring further improvement. While this gives us promising hints, but significant additional research and validation is required (especially on real-world datasets) in order to fully understand the tradeoffs and potential risks of this framework.

\textbf{Incorporating Cost Data for ROAS, and Optimization.} The current work focused on attribution based on media volume (e.g., impressions). A critical next step is to integrate media spend data, enabling the direct calculation of Return on Ad Spend (ROAS). Furthermore, exploring methods to apply marketing budget optimization techniques based on NNN's outputs would significantly enhance its practical applicability for marketers.

\textbf{Embeddings and Aggregation.} While NNN leverages pretrained model to generate embeddings, further investigation into optimal embedding techniques is warranted. This includes evaluating alternative pre-trained models (e.g., other language models besides the multilingual sentence encoder, different video models) \cite{t5, mt5} and developing effective representations for ad creatives, which remains relatively unexplored. Additionally, our current use of summation-based aggregation (SLaM) for embeddings (like individual search queries or ad views) may discard valuable distributional information. Future work should explore alternative aggregation methods—potentially revisiting analyses based on marginal distributions \cite{mulc2024compressing} or employing techniques like attention-based pooling—to better capture the richness of the underlying data.

\textbf{Model Architecture and Channel Modeling.} Several architectural refinements could improve NNN's performance and scope. First, extending the attention mechanism to explicitly model cross-geographical effects could capture spillover dynamics currently ignored. Second, implementing multi-head attention, rather than the current single-head approach, might allow different heads to specialize in distinct temporal patterns or channel interactions, potentially benefiting the multi-objective learning setup. Third, specific focus should be given to refining the modeling of key intermediary channels like Google Search, possibly incorporating more domain-specific structural priors. Fourth, while the current sales head incorporates saturation based on embedding direction, explicitly modeling volume-based saturation using established forms like the Hill function, particularly for marketing channel inputs, should be investigated. Finally, developing methods to utilize combined embeddings representing both ad content \textit{and} context simultaneously could lead to a more nuanced understanding of advertising effectiveness.

\textbf{Data Granularity.} Applying and evaluating NNN on more granular time scales, such as daily or even intra-day data where available, could reveal more dynamic marketing effects than weekly aggregation allows. NNN's design also permits exploration of non-uniform time granularities (e.g., daily marketing inputs predicting weekly sales), offering flexibility.

\textbf{Incorporating channels without embeddings} Currently NNN has only been tested with a two-channel framework (Search Ads and YouTube) using embedding data. This contrasts with MMMs which can model the effectiveness of dozens of channels at once. Further research is required to understand to what extent NNN can handle additional channels, especially channels that lack embedding data. 

\section{Conclusion}
This paper introduced NNN, a novel framework for Marketing Measurement leveraging the capabilities of neural networks, specifically Transformer architectures, and rich embedding representations for both paid media channels and organic signals like Google Search. Our approach utilizes the flexibility of neural networks to capture complex interactions and temporal dynamics, employs embeddings to integrate quantitative and qualitative channel information, and adopts common machine learning practices, such as optimization via L1 regularization and model selection based on out-of-distribution (OOD) predictive performance.

Through evaluations on both simulated and real-world datasets, we demonstrated NNN's potential. Our initial examples yielded plausible sales attribution estimates, on par with a baseline MMM. Additionally, when using NNN's framework of OOD performance for model assessment, we see considerable improvements in predictive accuracy relative to a baseline MMM.  Key aspects of our framework, such as the explicit modeling of intermediary channels like organic Search, offer an exciting potential mechanism to capture and attribute marketing effects potentially missed by conventional methods.  Moreover, the trained NNN model naturally lends itself as a flexible platform for diverse simulation-based analyses; for instance, evaluating the inferred effectiveness of specific creatives or keywords (i.e., creative analysis) can be performed directly within the framework, complementing traditional A/B testing. Finally, the model selection process, focused on tuning a single regularization parameter based on OOD performance, presents a potentially streamlined workflow for model development and selection.

While the results are promising, NNN is a novel research direction and should be viewed in context. It is not a direct replacement for currently established MMM frameworks, which benefit from decades of practical application, refinement, and established best practices. We see NNN as a complement rather than a replacement to MMM, with additional capabilities for channels with embedding data but a greater need for research and understanding of its uncertainties. We hope our methods and results inspire applications of neural networks in media measurement, live experiments of these techniques, and continued research in this area.

\section*{Acknowledgements}
Thank you to Nithya Mahadevan and Karima Zmerli for your steadfast support of this research and for invaluable advice along the way, and to Jennifer Steele for the early conversations about using Search as a proxy for consumer demand.  We would like to thank everyone in the Google Data Science community not explicitly mentioned above for their continued support and guidance.

\bibliographystyle{plain}
\bibliography{bib}

\section{Appendix}
\subsection{Glossary}
\label{sec:glossary}
\begin{itemize}
\item $\mathcal{C} := $ the set of all channels.
\item $C := |\mathcal{C}|$; the total number of channels.
% \item $\mathcal{G} :=$ the set of all geography (geo) regions.
\item $G := $ the number of geos.
\item $T := $ the total number of time-steps; For e.g., if we have 52 weeks of weekly data, the $T$ is 52.
\item $D := $ the maximum embedding dimension.
\item $\hat{\text{Sales}} := $ the predicted value of sales, or the main target, for all time steps and geos.
\item $\hat{\text{Search}} := $ the predicted value of search, or the main organic channel, for all time steps and geos.
\item $X :=$ the input data.
\item $F(X, \Theta) := $ NNN model, with parameters $\Theta$, that learns to predict the input data.
\item YT is shorthand for YouTube.
\item MMM generally refers to media mix models, and not our approach.
\end{itemize}

\subsection{Multiplicative Models}
Instead of additive models which sum the components of the different sales channels (Organic Search, Paid Search, YouTube, etc.), another formulation to allow synergy through a multiplicative model, which mixes the effect of all channels.  We train these models use the log-scale targets and adjust our prediction head to be

$$\sum_c V_t^c + \text{softplus}(\gamma^c_t) + M$$

where $M$ is the geo multiplier.  The sales targets during training are then $\log(\text{sales}_t)$.  Because we are training on log targets, we also scale all inputs of the model according to
$$v_{\text{scaled}} := \frac{v}{||v||_2} \cdot \log{||v||_2}$$
as a way to preserve the magnitude of the vector on the log-scale. We find that these models generally underfit the the analysis period, but do a good job at generalizing their capabilities into the out-of-distribution test period.

\subsection{L1 Regularization and Sparsity}
\begin{figure}[H]
    \centering
    \includegraphics[scale=.35]{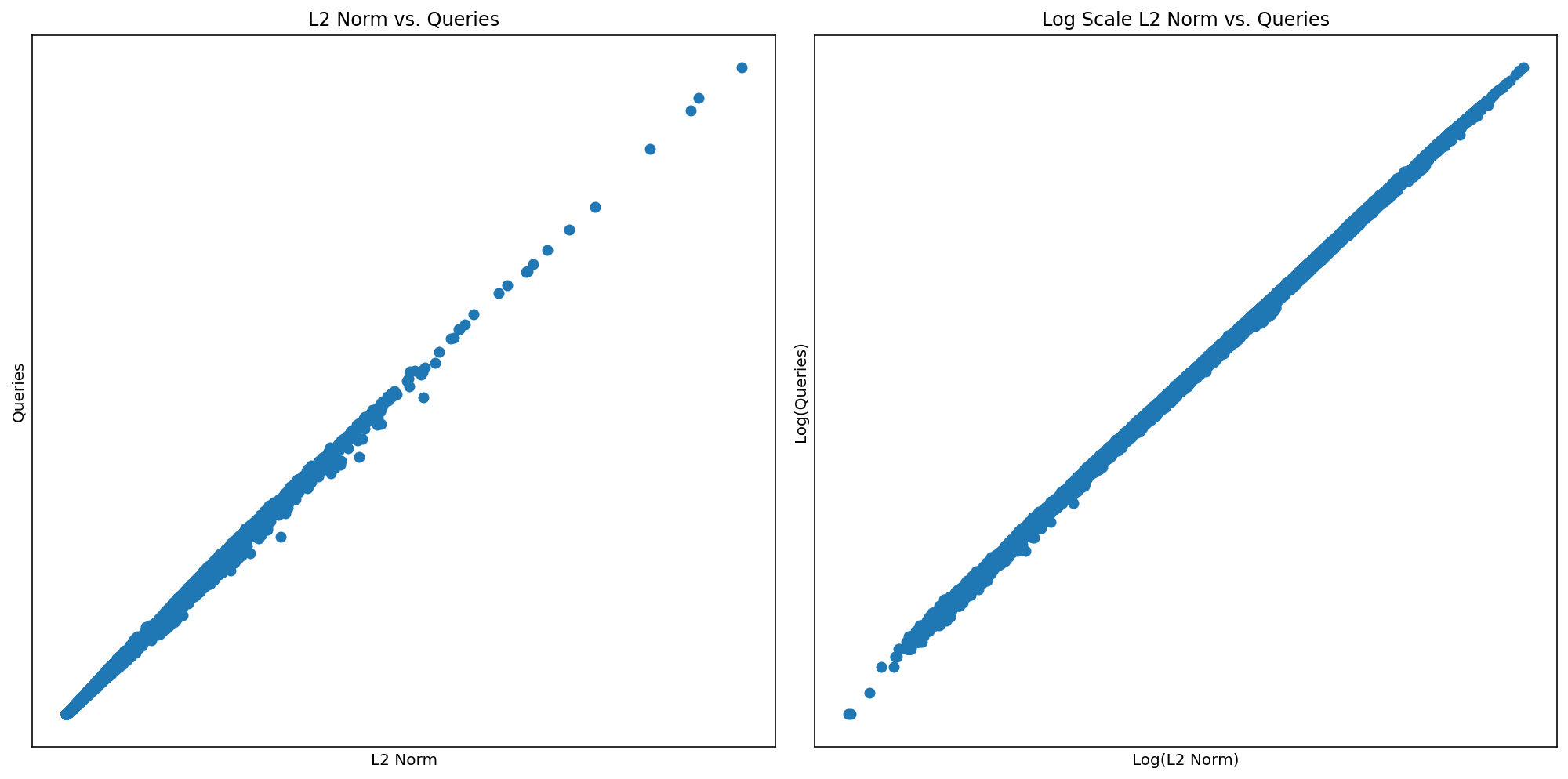}
    \caption{The L2 Norm of the search embedding is highly correlated (R=.99967) with the actual query volume, inspiring our decision to forgo storing the actual query volume and using the l2 as a substitute, thus simplifying the design.}
    \label{fig:norm_volume}
\end{figure}

\begin{figure} % Adjust width as needed
        \centering
        \includegraphics[width= .9 \textwidth]{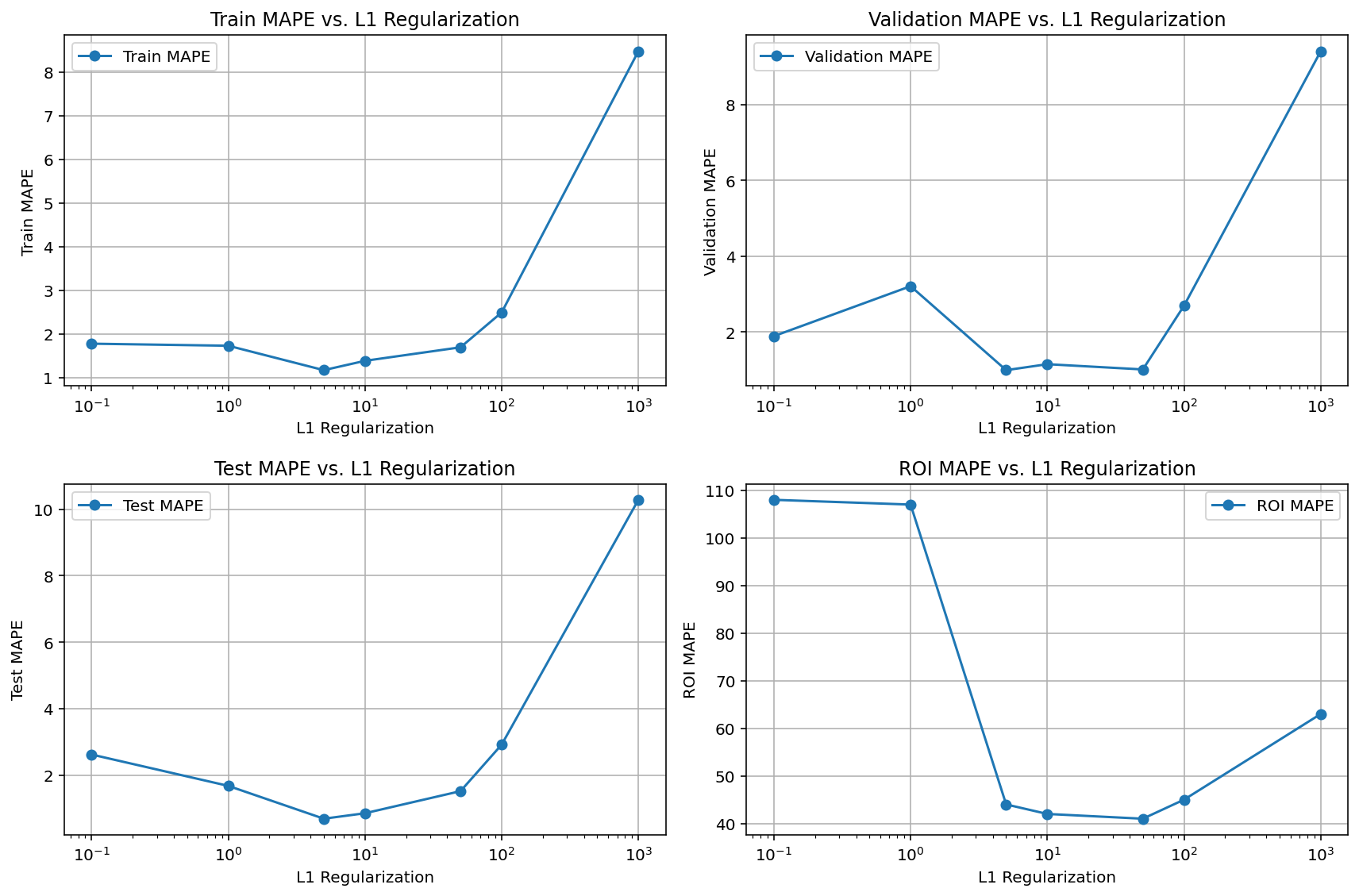} % Replace with your image file
        \caption{L1 Regularization vs Attribution Error and Predictive Error for the Low Variance Dataset.}
        \label{fig:low_var_l1}
\end{figure}

\subsection{Learned Temporal Attention}
\begin{figure}[H]
    \centering
    \includegraphics[scale=.5]{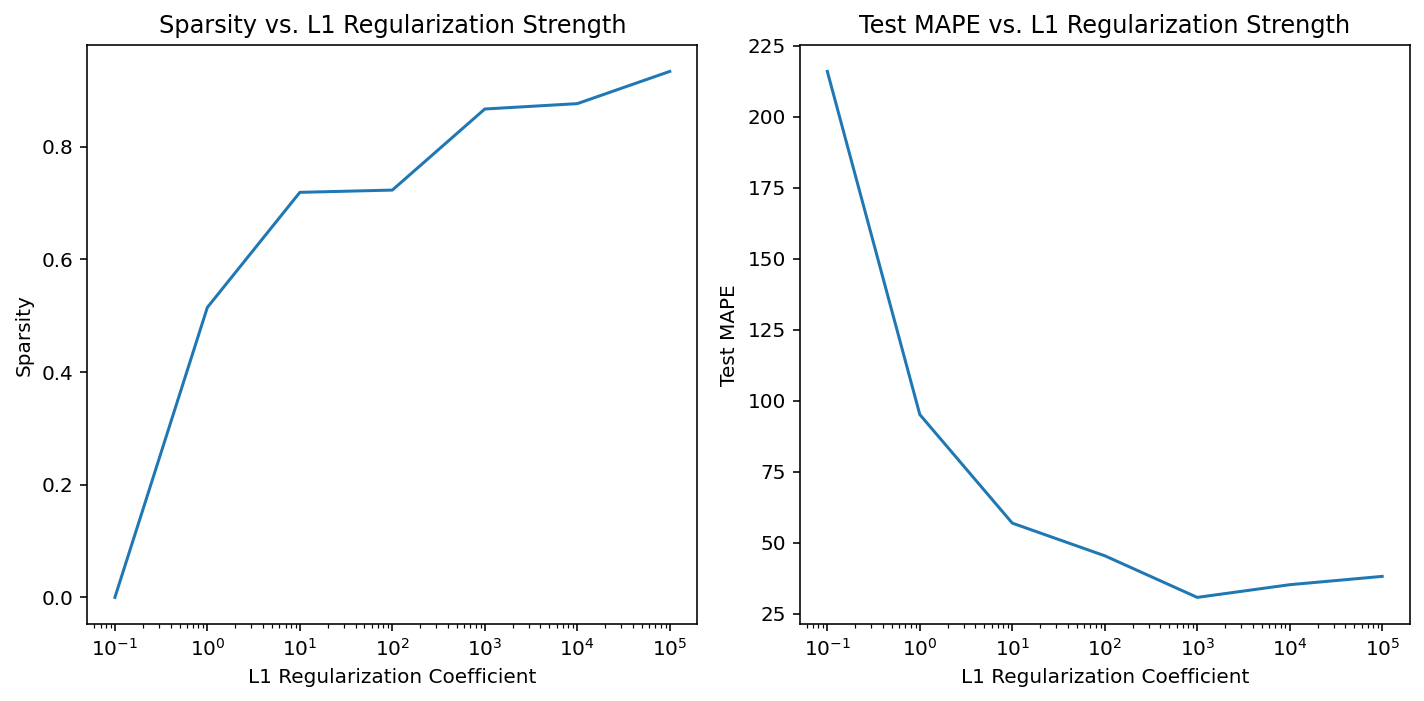}
    \caption{Regularization is an effective way to control the number of parameters (3M in our typical models) in the model that help reduce overfitting.  Plot were generated from a model trained using only the Sales target using inputs from Search, Search Ads, and YT Ads without channel mixing.}
    \label{fig:l1-reg}
\end{figure}

\subsection{Additional Model Plots}

\begin{figure}
    \centering
    \includegraphics[scale=.27]{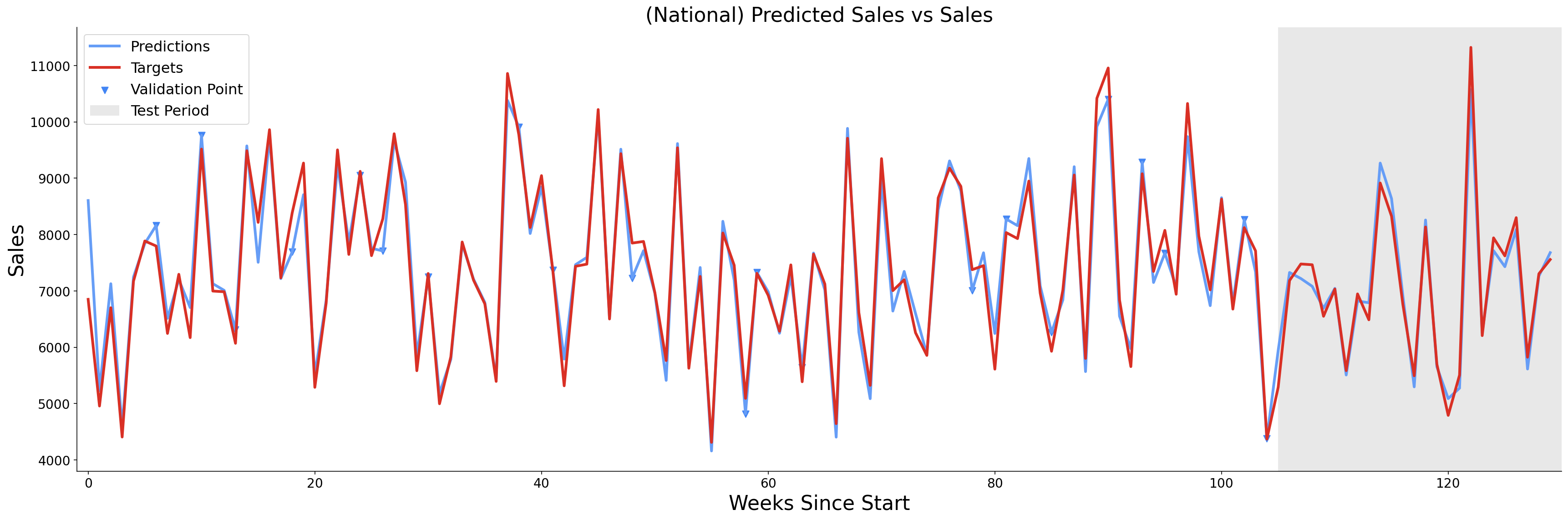}
    \caption{High variance, no channel mixing, after Search training, National plot.}
    \label{fig:high_variance_no_mix_search_natn}
\end{figure}

\begin{figure}
    \centering
    \includegraphics[scale=.36]{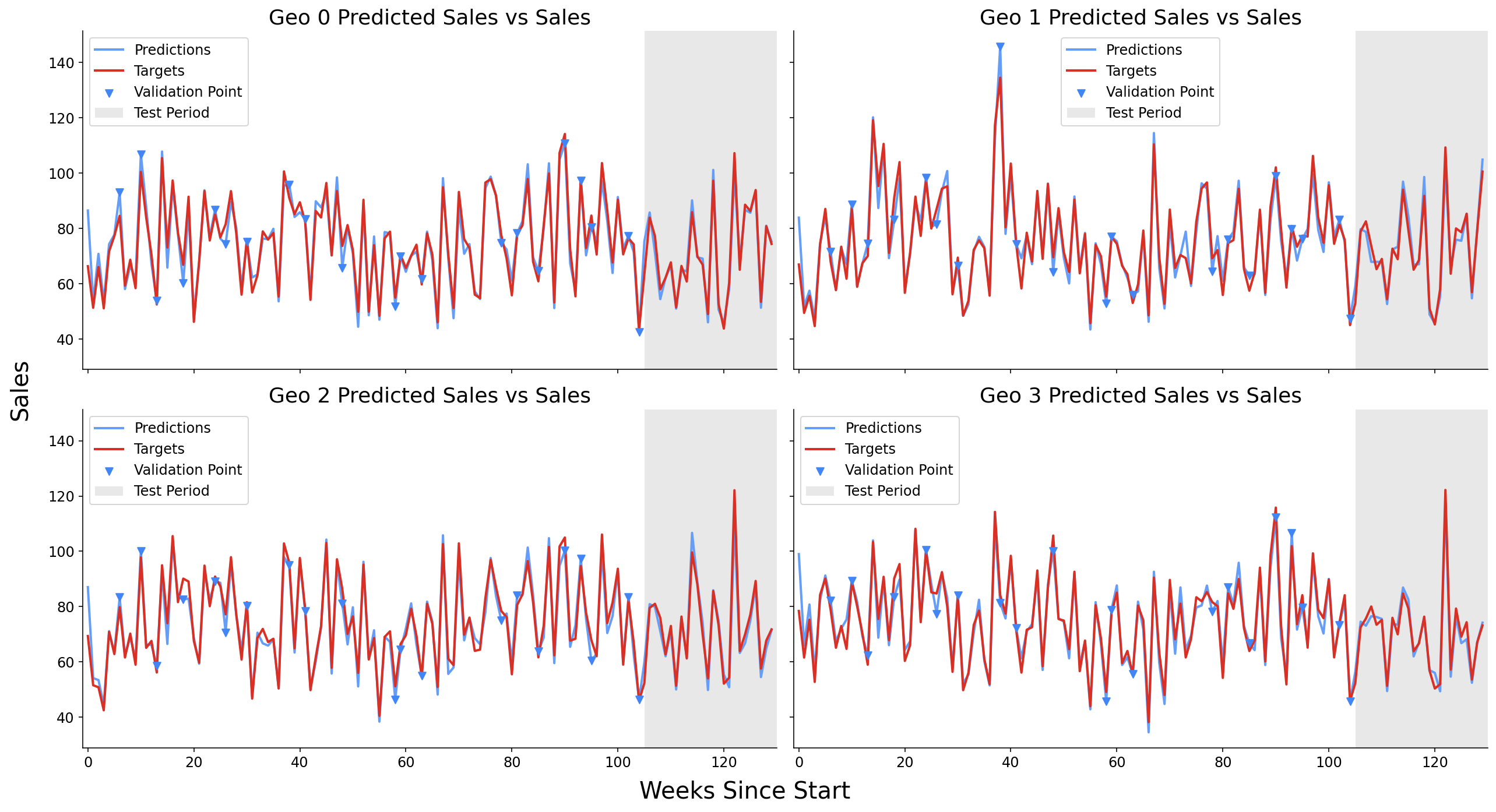}
    \caption{High variance, no channel mixing, after Search training, Geo plot.}
    \label{fig:high_variance_no_mix_search_geo}
\end{figure}

Figure \ref{fig:high_variance_no_mix_search_natn} shows the national predictions after search training with a balancing coefficient of .01.  For the same model, \ref{fig:high_variance_no_mix_search_geo} show the predictions for a random selection of 4 geos.  In here, we can see that the model predicts the spikes that occur in one geo but not the rest, highlighting considerable geo accuracy.

\subsection{Hyper Parameters}
\begin{table}[H]
    \centering
    \begin{tabular}{ll}
        \toprule
        \textbf{Hyperparameter} & \textbf{Value} \\
        \midrule
        Learning Rate & $10^{-4}$ \\
        Warm-up Steps & $100$ \\
        Training Steps & $5$k \\
        Gradient Noise & $10^{-5}$ \\
        Global Norm Clip & $1.0$\\
        Initial Learning Rate & $10^{-7}$ \\
        Decay Steps & $14$k \\
        L1 Regularization & $10$ \\
        Prediction Head Layers & $5$ \\
        Head Size & $64$ \\
        Transformer Blocks & $2$ \\
        Transformer Feed Forward Size & $512$ \\
        Balancing coefficient, $\alpha$  & .5 \\
        \bottomrule
    \end{tabular}
    \label{tab:hyperparameters}
      \caption{Default Hyperparameters}
\end{table}

\subsection{Pseudo Code}
\subsubsection{Factored Self-Attention}
\begin{lstlisting}[language=Python]
import flax
import flax.linen as nn
import jax.numpy as jnp


INDEX_MAPPING = flax.core.frozen_dict.FrozenDict(
    dict(sales=0, search=1, search_ads=2, yt=3))


class FactoredSelfAttention(nn.Module):
  """Implements self-attention factored into time and channel components.

  This module calculates self-attention weights based on two factors:
  1. Time: Attention scores are derived from relative time differences between
     positions using a learned MLP (`learned_time_scale`), optionally
     conditioned on the channel. This replaces standard dot-product attention 
     for the time dimension. A causal mask and a lookback window mask are
     applied.
  2. Channels: Attention scores between channels are learned directly via the
     `attn_scores_channels` parameter.

  The module can operate in two modes based on `channel_mixing`:
  - If `channel_mixing` is True, the time and channel attention weights are
    combined before aggregating the input values `x`. This allows attending
    across both time and channels simultaneously.
  - If `channel_mixing` is False, only the time attention weights are used
    to aggregate the input values `x` within each channel independently.
    Channel attention scores are still computed but not used for aggregation.

  Assumes input tensor `x` has shape (G, T, C, D), where G is geos,
  T is sequence length (time), C is number of channels, and D is feature
  dimension.

  Attributes:
    index_mapping: A mapping from channel names (str) to integer indices (int).
      Used primarily to determine the number of channels (C).
    lookback_window: The maximum number of past time steps to attend to.
      Attention scores for steps further back than this are masked out.
    temp: Temperature scaling factor applied before the softmax operations
      for both time and channel attention.
    channel_mixing: If True, combine time and channel attention weights before
      applying to the input `x`. If False, only apply time attention weights
      for aggregating `x`.
    attention_by_channel: If True, the learned time scale function receives
      channel identity information, allowing time attention scores to differ
      per channel. If False, time attention scores are calculated globally
      across channels."""
  index_mapping: flax.core.frozen_dict.FrozenDict[str, int]
  lookback_window: int = 52
  temp: float = 1.0
  channel_mixing: bool = False
  attention_by_channel: bool = True

  def setup(self):
    self.learned_time_scale = nn.Sequential(layers=[nn.Dense(128), nn.relu,
                                                    nn.Dense(1)])

    self.attn_scores_channels = self.param('attn_scores_channels',
                                           lambda key, shape: jnp.eye(shape),
                                           len(self.index_mapping))

  def __call__(self, x, pos_encoding, output_attention=False):
    # Assume input shape is (G, T, C, D)
    G, T, C, _ = x.shape

    # Time =====================================================================

    # Use the learned scale instead of dot-product attention
    time_deltas = (jnp.arange(T)[jnp.newaxis, :]
                   - jnp.arange(T)[:, jnp.newaxis]
                   ) / self.lookback_window  # (T, T)

    if self.attention_by_channel:
      attn_scores_time_input = jnp.expand_dims(time_deltas,
                                               axis=-1)  # (T, T, 1)
      attn_scores_time_input = jnp.broadcast_to(attn_scores_time_input,
                                                (G, T, T, C))
      attn_scores_time_input = jnp.expand_dims(attn_scores_time_input,
                                               axis=-1)  # (G, T, T, C,1)
      # One hot channel encoding
      channel_encoding = jnp.eye(C)  # (C, C)
      channel_encoding = jnp.expand_dims(channel_encoding,
                                         axis=(0, 1, 2))  # (1, 1, 1, C, C)
      channel_encoding = jnp.broadcast_to(channel_encoding, (G, T, T, C, C))

      attention_time_input = jnp.concatenate([attn_scores_time_input,
                                              channel_encoding],
                                             axis=-1)  # (G, T, T, C, C+1)
      attn_scores_time = self.learned_time_scale(
          attention_time_input)  # (G, T, T, C, 1)
      attn_scores_time = jnp.squeeze(attn_scores_time, axis=-1)  # (G, T, T, C)
      attn_scores_time = jnp.transpose(attn_scores_time,
                                       axes=(0, 1, 3, 2))  # (G, T, C, T)
    else:
      attn_scores_time_input = jnp.expand_dims(time_deltas, axis=-1)

      attn_scores_time = jnp.squeeze(self.learned_time_scale(
          attn_scores_time_input
          ), axis=-1)  # (T, T)

      attn_scores_time = jnp.expand_dims(attn_scores_time,
                                         axis=(0, 2))  # (1, T, 1, 1)
      attn_scores_time = jnp.broadcast_to(attn_scores_time, (G, T, C, T))

    # Set all entries that are outside the lookback_window to zero
    lookback_mask = jnp.triu(jnp.ones((T, T),
                                      dtype=jnp.float32),
                             k=-self.lookback_window + 1)

    # Mask out future positions
    mask = jnp.tril(jnp.ones((T, T),
                             dtype=jnp.float32))  # Lower triangular mask

    # Combine the masks: keep only the entries where both masks are 1
    mask = mask * lookback_mask

    mask = jnp.expand_dims(mask,
                           axis=(0, 2))  # Broadcast over batch and [heads (C) ]
    mask = jnp.broadcast_to(mask,
                            (G, T, C, T))  # Ensure matches the attention score

    # Mask out future positions and position not in window
    attn_scores_time = jnp.where(mask == 1,
                                 attn_scores_time,
                                 -jnp.inf)  # Mask out future positions

    # Softmax to compute attention weights
    attn_weights_time = nn.softmax(
        attn_scores_time / self.temp, axis=(-1,))  # (G, T, C, T)
    attn_weights_time = jnp.nan_to_num(
        attn_weights_time)  # In the case where there is nothing to attend to

    # Channels =================================================================

    # Learn an attention vector per each channel.
    attn_scores_channels = self.attn_scores_channels
    attn_scores_channels = jnp.expand_dims(attn_scores_channels, axis=(0, 1))
    attn_scores_channels = jnp.broadcast_to(attn_scores_channels, (G, T, C, C))

    # No need to mask out any channels
    attn_weights_channels = nn.softmax(attn_scores_channels / self.temp,
                                       axis=-1)  # (G, T, C, C)

    # Combine Channels and Time
    if self.channel_mixing:
      attn_weights = jnp.einsum('btcy,btcv->btcvy',
                                attn_weights_time,
                                attn_weights_channels)  # (G, T, C, C, T)

      attn_output = jnp.einsum('btcvy,byvd->btcd', attn_weights, x)
    else:
      attn_output = jnp.einsum('btcy,bycd->btcd', attn_weights_time, x)

    # Project the attention output back to the original dimensionality
    return self.out_proj(attn_output)
\end{lstlisting}

\subsubsection{MLPResnet}
\begin{lstlisting}[language=Python]
class MLPResnet(nn.Module):
  """Standard MLP Resnet model used in Compressing Search with LMs.

  https://arxiv.org/abs/2407.00085

  The input is expected to have shape (*other_dims, D) and the model works
  by first projecting it to layer_size and then applying the resnet block
  n_layers times. The output is then projected back to the original dimension.
  (or output_dim if specified)
  """
  n_layers: int
  layer_size: int = 64
  project_back: bool = True
  output_dim: int| None = None

  @nn.compact
  def __call__(self, x):
    dim_size = x.shape[-1]
    x = nn.Dense(self.layer_size)(x)
    for _ in range(self.n_layers):
      x += nn.relu(nn.Dense(self.layer_size)(x))
    if self.project_back:
      return nn.Dense(dim_size)(x)
    elif self.output_dim is not None:
      return nn.Dense(self.output_dim)(x)
    return x
\end{lstlisting}

\subsection{Transformer Layer}
\begin{lstlisting}[language=Python]
class TransformerLayer(nn.Module):
  """Transformer layer using custom multi-head attention across channels."""
  d_model: int
  d_ff: int
  channels: int
  index_mapping: flax.core.frozen_dict.FrozenDict[str, int]
  channel_mixing: bool

  def setup(self):
    self.self_attn = FactoredSelfAttention(index_mapping=self.index_mapping,
                                           channel_mixing=self.channel_mixing)
    self.ffns = [nn.Sequential([
        nn.Dense(self.d_ff),
        nn.relu,
        nn.Dense(self.d_model)
    ]) for _ in range(self.channels)]

  def __call__(self, x, pos_encoding):
    attn_output = self.self_attn(x,)  # (B, T, C, D)
    x = x + attn_output

    # Feed-forward network
    # Non linear ffn for each channel
    ffn_channels = []
    for i in range(self.channels):
      projected = self.ffns[i](x[:, :, i, :])  # (B, T, D')
      ffn_channels.append(jnp.expand_dims(projected, axis=2))  # (B, T, 1, D')

    # Concatenate the projected channels along the channel dimension (axis=2)
    ffn_output = jnp.concatenate(ffn_channels, axis=2)  # (B, T, C, D')

    # Add with the residual connection again
    return x + ffn_output
\end{lstlisting}

\subsection{Sales Prediction Head}
\begin{lstlisting}[language=Python]
class SalesHead(nn.Module):
  """Sales Head.

  Takes a D-Dimensional Vector, and learns a non-linear transformation of the
  normed version.  Passes the final output through a sigmoid, then scaled by
  the original l2 norm.
  """
  n_layers: int
  layer_size: int = 64
  output_dim: int | None = None
  log_scale: bool = True

  @nn.compact
  def __call__(self, x, geo, use_geo=True):
    """Sales Head with Multipliers.

    Args:
      x: A JAX array with shape (G, T, D).
      geo: A JAX array with shape (G, T, G) containing the geo one-hot.
      use_geo: A boolean indicating whether to use the geo multiplier.

    Returns:
      A JAX array with shape (G, T, D)
    """
    norm = l2_norm(x)
    x /= norm
    if use_geo:
      x = jnp.concatenate([x, geo], axis=-1)  # (G, T, D + G)

    x = MLPResnet(n_layers=self.n_layers,
                  layer_size=self.layer_size,
                  output_dim=self.output_dim,
                  project_back=False)(x)
    if use_geo:
      if not self.log_scale:
        geo_mult = nn.Dense(1, use_bias=False,
                            kernel_init=nn.initializers.ones)(geo)  # Geo multiplier
      else:
        geo_mult = nn.Dense(1, use_bias=False,)(geo)
    else:
      if self.log_scale:
        geo_mult = 0.
      else:
        geo_mult = 1.

    if not self.log_scale:
      geo_mult = jnp.abs(geo_mult)

    if self.log_scale:
      return norm -  nn.softplus(x) + geo_mult
    else:
      return norm * nn.sigmoid(x) * geo_mult
      
      
class SalesMultiChannelResnetHead(nn.Module):
  """Sales Model Head for Multi-Channel Inputs."""
  n_layers: int
  output_dim: int
  log_scale: bool
  layer_size: int = 64
  channels_to_use: tuple[int, ...] = (INDEX_MAPPING['search'],)

  @nn.compact
  def __call__(self, x, geos):
    outputs = []
    for c in self.channels_to_use:
      if c == INDEX_MAPPING['search']:
        use_geo = True
      else:
        use_geo = False
      temp = SalesHead(self.n_layers,
                      layer_size=self.layer_size,
                      output_dim=self.output_dim,
                      log_scale=self.log_scale)(
                                  x[:, :, c, :], geos, use_geo=use_geo)  # (B, T, D)
      temp = jnp.expand_dims(temp, axis=2)  # (B , T, 1, D)
      outputs.append(temp)
    outputs = jnp.concatenate(outputs, axis=2)  # (B, T, C, D)
    return outputs.sum(axis=2)  # (B, T, D)
\end{lstlisting}

\end{document}